\pdfoutput=1

\documentclass[11pt]{article}

\usepackage[]{acl}

\usepackage{times}
\usepackage{latexsym}

\usepackage[T1]{fontenc}

\usepackage[utf8]{inputenc}

\usepackage{microtype}
\usepackage{subcaption}
\usepackage{graphicx}
\usepackage{booktabs}
\usepackage{amsmath}
\usepackage{amsfonts}
\usepackage{paralist}
\usepackage{algorithm}
\usepackage{algpseudocode}
\usepackage{mathrsfs}
\usepackage{enumitem}
\usepackage{tabu}
\usepackage{multirow}
\usepackage{amsmath}
\usepackage{amssymb}
\usepackage{amsthm}
\usepackage{amsfonts}
\usepackage{algorithmicx}
\usepackage{algorithm}
\usepackage{algc}
\usepackage{algcompatible}
\usepackage{algpseudocode}

\newtheorem{thm}{Theorem}

\newtheorem*{property}{Property}

\newcommand{\mcal}{\mathcal}

\newcommand\given[1][]{\:#1\vert\:}

\title{On the Paradox of Learning to Reason from Data}
\author{Honghua Zhang, Liunian Harold Li, Tao Meng, \\
  \textbf{Kai-Wei Chang, Guy Van den Broeck} \\
  University of California, Los Angeles
}

\begin{document}

\maketitle
\begin{abstract}
Logical reasoning is needed in a wide range of NLP tasks. Can a BERT model be trained end-to-end to solve logical reasoning problems presented in natural language? We attempt to answer this question in a \emph{confined problem space} where there exists a set of parameters that \emph{perfectly simulates} logical reasoning. We make observations that seem to contradict each other: BERT attains near-perfect accuracy on in-distribution test examples while failing to generalize to other data distributions over the exact \emph{same} problem space. Our study provides an explanation for this paradox: instead of learning to emulate the correct reasoning function, BERT has in fact learned \emph{statistical features} that inherently exist in logical reasoning problems. We also show that it is infeasible to jointly remove statistical features from data, illustrating the difficulty of learning to reason in general. Our result naturally extends to other neural models and unveils the fundamental difference between learning to reason and learning to achieve high performance on NLP benchmarks using statistical features. 
\end{abstract}

\section{Introduction}
\label{sec:intro}
Logical reasoning is needed in a wide range of NLP tasks including natural language inference~(NLI)~\citep{williams2018mnli,bowman2015snli}, question answering~(QA)~\citep{rajpurkar2016squad,yang2018hotpotqa} and common-sense reasoning~\citep{zellers2018swag,talmor2019commonsenseQA}. The ability to draw conclusions based on given facts and rules, is essential to solving these tasks.\footnote{A.k.a., deductive reasoning. In this paper, we do not consider inductive reasoning, where rules need to be learned.} Though NLP models, empowered by the Transformer neural architecture~\citep{vaswani2017transformer}, can achieve high performance on task-specific datasets~\citep{devlin2019bert}, it is unclear whether they are ``reasoning'' over the input following the rules of logic. A research question naturally arises: \emph{can neural networks be trained to conduct logical reasoning presented in natural language?}

\begin{figure}[t]
    \centering
      \includegraphics[width=0.9\linewidth]{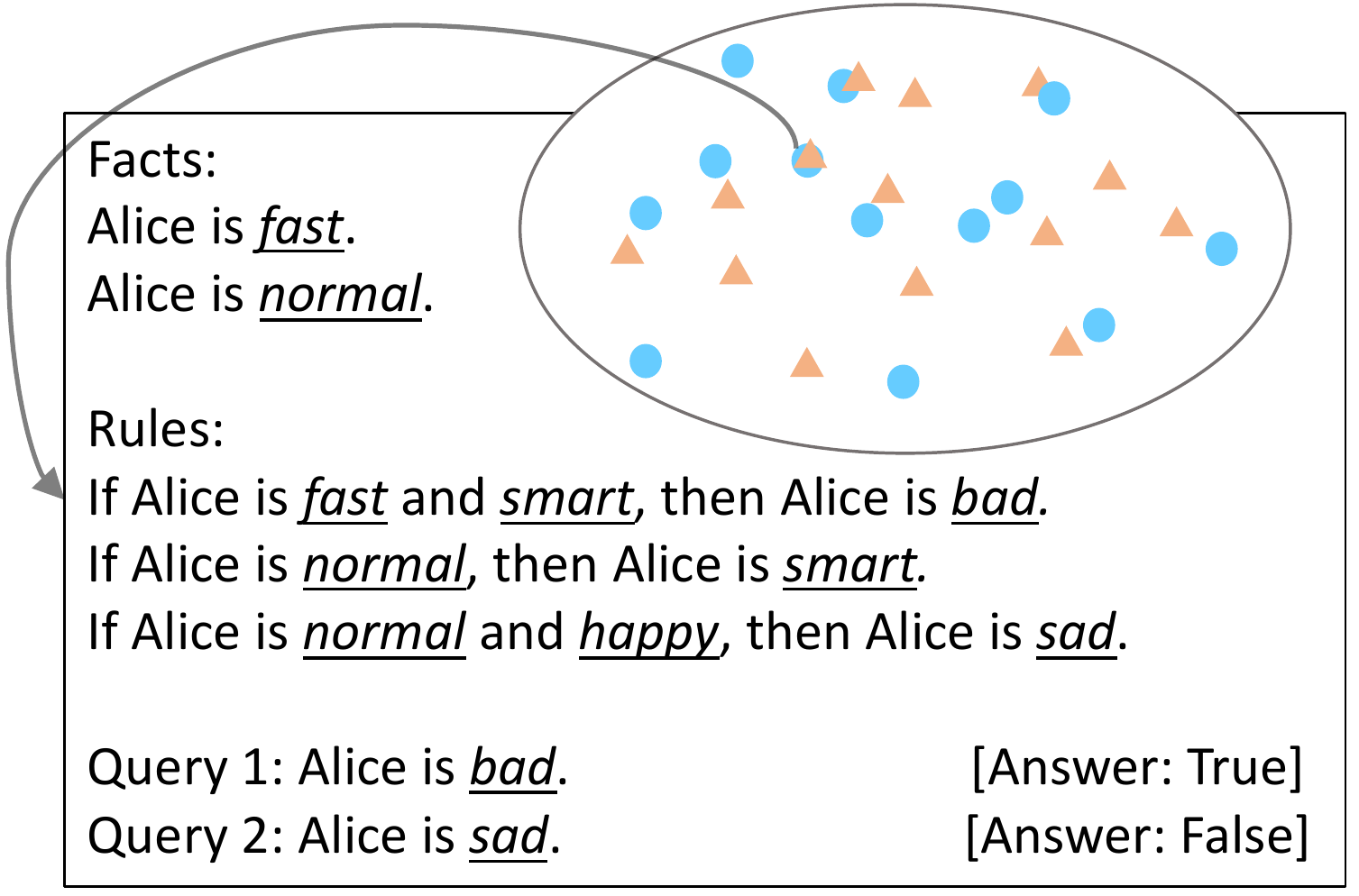}
    \caption{Problem setting: a confined problem space consisting of logical reasoning problems; dots and triangles denote examples sampled from different distributions over the same problem space.}
    \label{fig:reasoning_examples}
\end{figure}

Following prior work, we attempt to answer this question by training and testing a neural model~(e.g. BERT~\citep{devlin2019bert}) on a \textbf{confined problem space}~(SimpleLogic, see Sec.~\ref{sec:simplelogic} and Fig.~\ref{fig:reasoning_examples}) consisting of logical reasoning problems written in English~\citep{johnson2017clevr,sinha2019clutrr,clark2020ruletaker}. Yet, we observe evidences that seemingly lead to a contradiction.

On the one hand, the following evidences seem to imply that \textcolor{black}{\textit{neural models can learn to reason}}:

\indent \textit{E1:} Logical reasoning problems in the problem space are self-contained: they have no language variance and require no prior knowledge.

\indent \textit{E2:} We show that theoretically, the BERT model has enough capacity to represent the correct reasoning function~(Sec~\ref{subsec:construction}).

\indent \textit{E3:} The BERT model can be trained to achieve near-perfect test accuracy on a data distribution covering the whole problem space.
    
With these evidences (and with even fewer evidences), it has been claimed that neural models can learn to reliably emulate the correct reasoning function~\citep{clark2020ruletaker, talmor2020leap}.

On the other hand, however, we observe a seemingly contradictory phenomenon: the models attaining near-perfect accuracy on one data distribution do not generalize to other distributions within \emph{the same problem space}. Since the correct reasoning function does not change across data distributions, it follows that \textcolor{black}{\textit{the model has not learned to reason}}.

The paradox lies in that if a neural model \emph{has} learned reasoning, it should not exhibit such a generalization failure; if the model \emph{has not} learned reasoning, it is baffling how it manages to achieve near-perfect test accuracy on a training distribution that covers the entire problem space. Note that what we observed is not a common case of out-of-distribution~(OOD) generalization failure: (1)~the problem space is confined and simple~(see \textit{E1,E2}); (2)~the correct reasoning function is invariant across data distributions; (3)~the training distribution covers the whole problem space. On the contrary, discussions on OOD generalization often involve open problem space~\citep{lin2019reasoning, gontier2020measuring, wald2021calibration} and domain/concept mismatch between training and testing distribution~\citep{yin2021broaden, koh2021wilds}.

Upon further investigation, we provide an explanation for this paradox: the model attaining high accuracy \textbf{only} on in-distribution test examples \textbf{has not learned to reason}. In fact, the model has learned to use \emph{statistical features} in logical reasoning problems to make predictions rather than to emulate the correct reasoning function.

Our first observation is that even the simplest statistic of a reasoning problem can give away significant information about the true label~(Sec.\ref{subsec:statistical_feature_exists}):~for example, by only looking at the number of rules in a reasoning problem, we can predict the correct label better than a random guess. Unlike dataset biases/artifacts identified in typical NLP datasets, which are often due to biases in the dataset collection/annotation process~\citep{gururangan2018annotation, clark2019dont, he2019unlearn}, statistical features \textbf{inherently} exist in reasoning problems and are not specific to certain data distributions. We show that statistical features can hinder model generalization performance; moreover, we argue that there are potentially countless statistical features and it is computationally expensive to jointly remove them from training distributions.

Our study establishes the dilemma of learning to reason from data: on the one hand, when a model is trained to learn a task from data, it always tends to learn statistical patterns, which inherently exist in reasoning examples; on the other hand, however, the rules of logic never rely on statistical patterns to conduct reasoning. Since it is difficult to construct a logical reasoning dataset that contains no statistical features, it follows that learning to reason from data is difficult. 

Our findings unveil the fundamental difference between ``learning to reason'' and ``learning to solve a typical NLP task.'' For most NLP tasks, one of the major goal for a neural model is to learn statistical patterns: for example, in sentiment analysis~\cite{imdb}, a model is \emph{expected} to learn the strong correlation between the occurrence of the word ``happy'' and the positive sentiment. However, for logical reasoning, even though numerous statistical features inherently exist, models should not be utilizing them to make predictions. Caution should be taken when we seek to train neural models end-to-end to solve NLP tasks that involve both logical reasoning and prior knowledge and are presented with language variance~\citep{welleck2021naturalproofs, yu2020reclor}, which could potentially lead to even stronger statistical features, echoing the findings of~\citet{elazar2021back, mccoy2019right}.

\section{SimpleLogic: A Simple Problem Space for Logical Reasoning}
\label{sec:simplelogic}

We define \emph{SimpleLogic}, a class of logical reasoning problems based on propositional logic. We use SimpleLogic as a controlled testbed for testing neural models' ability to conduct logical reasoning. 

SimpleLogic only contains deductive reasoning examples. To simplify the problem, we remove language variance by representing the reasoning problems in a templated language and limit their complexity (e.g., examples have limited input lengths, number of predicates, and reasoning depths). 

Solving SimpleLogic does not require significant model capacity. We show that a popular pre-trained language model BERT \cite{devlin2019bert}\footnote{BERT is one of the most popular language model backbones for NLP downstream models. In this paper, we use BERT as a running example and our conclusion can be naturally extended to other Transformer-based NLP models.} has more than enough model capacity to solve SimpleLogic by constructing a parameterization of BERT that solves SimpleLogic with 100\% accuracy~(Sec.~\ref{subsec:construction}).

\subsection{Problem Space Definition}
\label{subsec:simplelogic}
Before defining SimpleLogic, we introduce some basics for propositional logic. In general, reasoning in propositional logic is NP-complete; hence, we only consider propositional reasoning with \emph{definite clauses}. A definite clause in propositional logic is a \emph{rule} of the form $A_1 \land A_2 \land \cdots \land A_n \rightarrow B$, where $A_i$s and $B$ are \emph{predicates} that take values in ``True'' or ``False''; we refer to the left hand side of a rule as its \emph{body} and the right hand side as its \emph{head}. In particular, a definite clause is called a \emph{fact} if its body is empty (i.e. $n = 0$). A \emph{propositional theory} (with only definite clauses) $T$ is a set of rules and facts, and we say a predicate $Q$ \emph{can be proved} from $T$ if either (1) $Q$ is given in $T$ as a fact or (2) $A_1 \land \cdots \land A_n \rightarrow Q$ is given in $T$ as a rule where $A_i$s can be proved. 

Each example in SimpleLogic is a propositional reasoning problem that only involves definite clauses. In particular, each example is a tuple (\emph{facts}, \emph{rules}, \emph{query}, \emph{label}) where (1) \emph{facts} is a list of predicates that are known to be True, (2) \emph{rules} is a list of rules represented as definite clauses, (3) \emph{query} is a single predicate, and (4) \emph{label} is either True or False, denoting whether the query predicate can be proved from \emph{facts} and \emph{rules}. Figure~\ref{fig:reasoning_examples} shows such an example.
Furthermore, we enforce some simple constraints to control the difficulty of the problems. For each example in SimpleLogic, we require that:
\begin{compactitem}
    \item the number of predicates (\#pred) that appear in facts, rules and query ranges from $5$ to $30$, and all predicates are sampled from a fixed vocabulary containing $150$ adjectives such as ``happy'' and ``complicated''; note that the predicates in SimpleLogic have \textbf{no} semantics;
    \item the number of rules (\#rule) ranges from 0 to $4 \times $ \#pred, and the body of each rule contains $1$ to $3$ predicates; i.e. $A_1 \land \ldots \land A_n \rightarrow B$ with $n > 3$ is not allowed; 
    \item the number of facts (\#fact) ranges from $1$ to \#pred;
    \item the reasoning depth\footnote{For a query with label \emph{True}, its reasoning depth is given by the depth of the shallowest proof tree; for a query with label \emph{False}, its reasoning depth is the maximum depth of the shallowest failing branch in all \emph{possible} proof trees.} required to solve an example ranges from $0$ to $6$.
\end{compactitem}
We use a simple template to encode examples in SimpleLogic as natural language input. For example, we use ``\emph{Alice is X.}'' to represent the fact that $X$ is True; we use ``\emph{A and B, C.}'' to represent the rule $A \land B \rightarrow C$; we use ``\emph{Query: Alice is Q.}'' to represent the query predicate $Q$. Then we concatenate \emph{facts}, \emph{rules} and \emph{query} as \emph{[CLS] facts. rules [SEP] query [SEP]} and supplement it to BERT to predict the correct \emph{label}.

\begin{figure}
  \centering
  \includegraphics[width=0.9\linewidth]{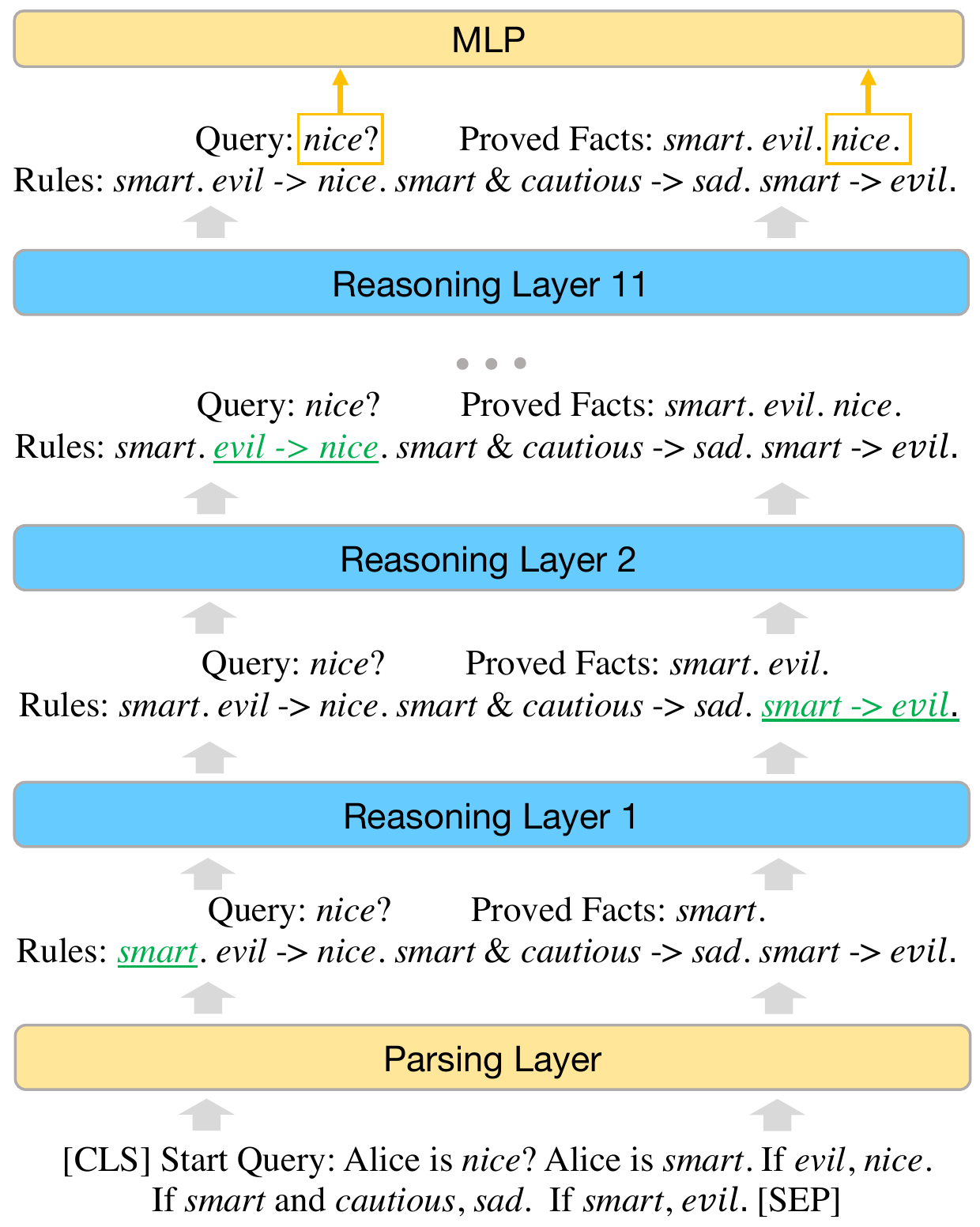}
    \caption{A BERT-base model that simulates the forward-chaining algorithm. The first layer parses text input into the desired format. Each reasoning layer performs one step of forward-chaining, adding some predicates to the Proved Facts, and the rules being used are underlined in green; e.g. Reasoning Layer 2 use the rule ``smart $\rightarrow$ evil'' to prove the predicate \emph{evil}. }
    \label{fig:construction}
\end{figure}

\subsection{BERT Has Enough Capacity to Solve SimpleLogic}
\label{subsec:construction}

In the following, we show that BERT has enough capacity to solve all examples in SimpleLogic. In particular, we explicitly construct a parameterization for BERT such that the fixed-parameter model solves all problem instances in SimlpleLogic. Note that we only prove the existence of such a parameterization, but do not discuss whether such a parameterization can be learned from sampled data until Sec. \ref{sec:story1}.

\begin{thm}
\label{thm:construction}
For BERT with $n$ layers, there exists a set of parameters such that the model can correctly solve any reasoning problem in SimpleLogic that requires $\leq n-2$ steps of reasoning. 
\end{thm}

We prove this theorem by construction; in particular, we construct a fixed set of parameters for BERT to simulate the forward-chaining algorithm. 

Here we show a sketch of the proof, and refer readers to Appendix~\ref{appendix:construction} for the full proof. As illustrated in Figure~\ref{fig:construction}, our construction solves a logical reasoning example in a layer-by-layer fashion. The 1st layer of BERT parses the input sequence into the desired format. Layer 2 to layer 10 are responsible for simulating the forward chaining algorithm: each layer performs one step of reasoning, updating the True/False label for predicates. The last layer also performs one step of reasoning, while implicitly checking if the query predicate has been proven and propagating the result to the first token. The parameters are the same across all layers except for the Parsing Layer (1st layer).

We implemented the construction in PyTorch, following the exact architecture of the BERT-base model. The ``constructed BERT'' solves all the problems in SimpleLogic of reasoning depth $\leq 10$ with 100\% accuracy, using only a small proportion of the parameters.\footnote{Code available at \url{https://github.com/joshuacnf/paradox-learning2reason}.}

\section{BERT Fails to Learn to Solve SimpleLogic}
\label{sec:story1}
In this section, we study whether it is possible to train a neural model (e.g., BERT) to reason on SimpleLogic. We follow prior work~\citep{clark2020ruletaker} to randomly sample examples from the problem space and train BERT on a large amount of sampled data. We consider two natural ways to sample data from SimpleLogic and expect that if a model has learned to reason, the model should be able to solve examples generated by any sampling methods.

\begin{figure}
  \centering
  \includegraphics[width=1.0\linewidth]{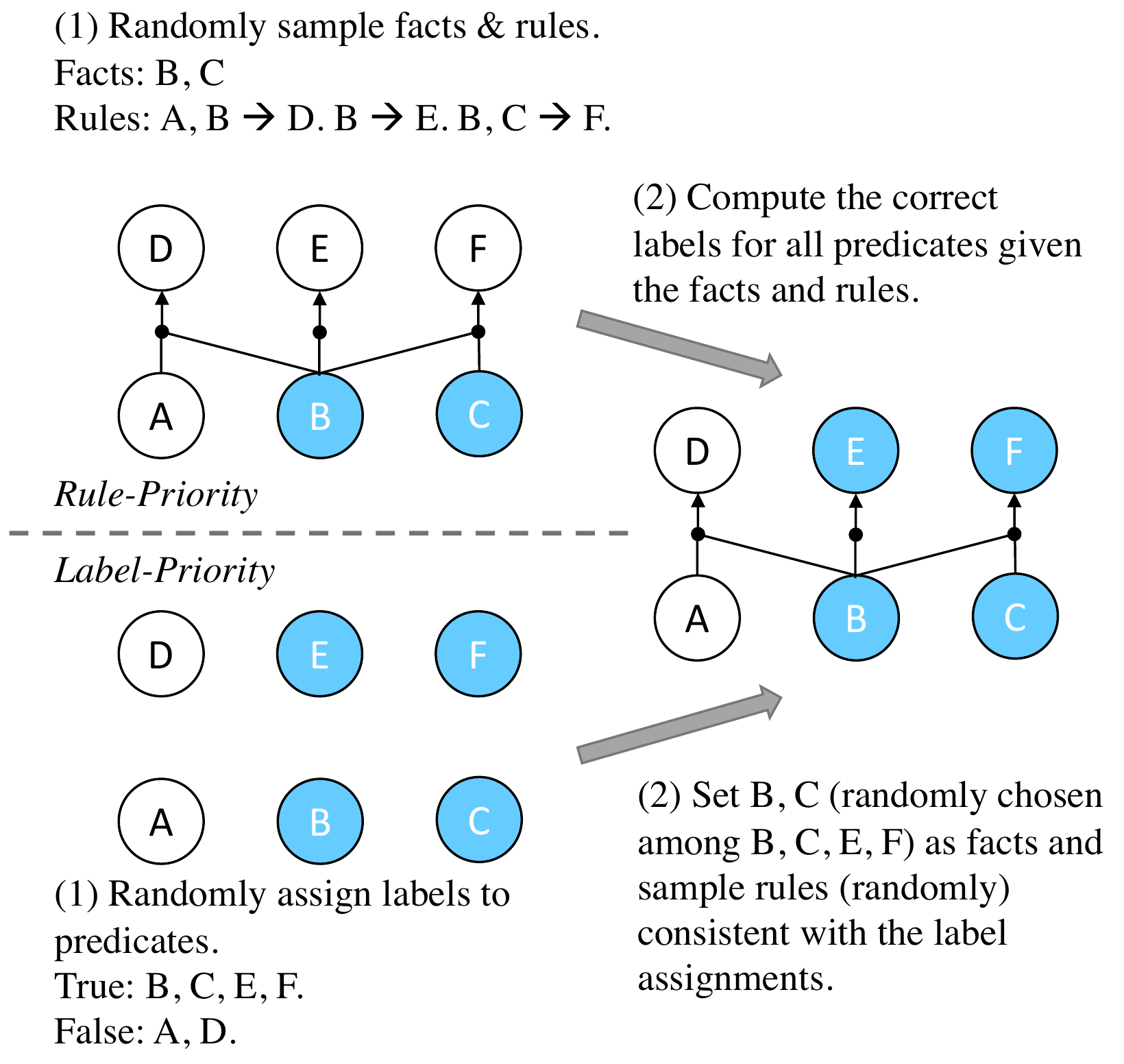}
    \caption{An illustration of a logical reasoning problem (right) in SimpleLogic being sampled by Rule-Priority~(RP) and Label-Priority~(LP), respectively. Predicates with label \emph{True} are denoted by filled circles.}
    \label{fig:RPLP}
\end{figure}

\subsection{Sampling Examples from SimpleLogic}
\label{subsec:RP_LP}
When sampling examples from a finite domain, one naive approach is to uniformly sample from the domain. However, uniform sampling is \textbf{not desirable}: by computation, it is easy to show that over $99.99\%$ of the examples generated by uniform sampling have $30$ predicates and $120$ rules. This is a serious problem in terms of \emph{coverage}: we expect a reasonable dataset to cover reasoning examples of different \#pred, \#fact and \#rule. Hence, we instead consider the following two intuitive ways of sampling examples:

\paragraph{Rule-Priority (RP).} To solve the major issue with uniform sampling, in Rule-Priority, we first randomly sample \#pred, \#fact and \#rule uniformly at random from $[5, 30]$, $[1, \text{\#pred}]$ and $[1, 4\times\text{\#pred}]$ respectively, ensuring that all three aspects are covered by a non-trivial number of examples. Then, we randomly sample some predicates, facts and rules based on the given \#pred, \#rule and \#fact. The query is also randomly sampled, and its label is computed by forward-chaining based on the given facts and rules.

\paragraph{Lable-Priority (LP).} In Rule-Priority, we first randomly generate rules and facts, which then determines the label for each predicate. In Label-Priority (LP), we consider generating examples in the ``reversed'' order: we first randomly assign a True/False label to each predicate and then randomly sample some rules and facts that are \emph{consistent} with the pre-assigned labels.

Figure~\ref{fig:RPLP} shows an example illustrating the two sampling methods. Both LP and RP are very general, covering the whole problem space. We refer readers to the Appendix for further details on the sampling algorithms.

\subsection{BERT Trained on Randomly Sampled Data Cannot Generalize}
\label{subsec:story1_result}
Following the two sampling regimes described above, we randomly sample two sets of examples from SimpleLogic: for each reasoning depth from $0$ to $6$, we sample $80k$ examples from SimpleLogic via algorithm RP (LP) and aggregate them as dataset RP (LP), which contains $560k$ examples in total. We then split it as training/validation/test set.
We train a BERT-base model \citep{devlin2019bert} on RP and LP, respectively. We train for 20 epochs with a learning rate of $4\times10^{-5}$, a warm-up ratio of $0.05$, and a batch size of 64. Training takes less than 2 days on 4 NVIDIA 1080Ti / 2080Ti GPUs with 12Gb GPU memory.

\paragraph{BERT performs well on the training distribution.}
The first and last rows of Table~\ref{table:RP_LP} show the test accuracy when the test and train examples are sampled by the same algorithm (e.g., for row~1, the model is trained on the training set of RP and tested on the test set of RP). In such scenarios, the models can achieve near-perfect performance similar to the observations in prior work \citep{clark2020ruletaker}. Both sampling algorithms are general in the sense that every instance in SimpleLogic has a positive probability to be sampled; hence, the intuition is that the model has learned to emulate the correct reasoning function.

\paragraph{BERT fails to generalize.}
However, at the same time, we observe a rather counter-intuitive finding: the test accuracy drops significantly when the train and test examples are sampled via different algorithms. Specifically, as shown in the second and third rows of Table~\ref{table:RP_LP}, the BERT model trained on RP fails drastically on LP, and vice versa. 
Since the correct reasoning function does not change across different data distributions, this generalization failure indicates BERT is has not learned to conduct logical reasoning.
A subsequent question naturally arise: is this simply because LP and RP are complementary? Can the model learn to reason if we train the model on data sampled by both algorithms?

\paragraph{Training on both RP and LP is not enough.}
We train BERT on the mixture of RP and LP, and BERT again achieves nearly perfect test accuracy. Can we now conclude that BERT has learned to approximate the correct reasoning function?  We slightly tweak the sampling algorithm of LP by increasing the expected number of alternative proof trees to generate LP$^*$. Unfortunately, we observe that the model performance again drops significantly on LP$^*$ (Table~\ref{table:RP_LP_2}). 
Such a result resembles what we observed in Table~\ref{table:RP_LP}, even when we are enriching our training distribution with different sampling methods. In fact, we find \emph{no evidence} that consistently enriching the training distribution will bring a transformative change such that the model can learn to reason.

\begin{table}[]
    \setlength{\tabcolsep}{4pt}
    \centering
    
    {\footnotesize
    \resizebox{\linewidth}{!}{
    \begin{tabular}{cc| c c c c c c c } 
        \toprule
        Train & Test & $0$ & $1$ & $2$ & $3$ & $4$ & $5$ & $6$  \\ 
        \midrule
        \multirow{2}{*}{RP} & RP & 99.9 & 99.8 & 99.7 & 99.3 & 98.3 & 97.5 & 95.5\\
         & LP & 99.8 & 99.8 & 99.3 & 96.0 & \textcolor{red}{90.4} & \textcolor{red}{75.0} & \textcolor{red}{57.3}\\
        \midrule
        \multirow{2}{*}{LP} 
            & RP & 97.3 & \color{red}{66.9} & \color{red}{53.0} & \color{red}{54.2} & \color{red}{59.5} & \color{red}{65.6} & \color{red}{69.2}  \\
        & LP & 100.0 & 100.0 & 99.9 & 99.9 & 99.7 & 99.7 & 99.0 \\
        \bottomrule
    \end{tabular}
    }}
    \caption{
    Test accuracy on LP/RP for the BERT model trained on LP/RP; the accuracy is shown for test examples with reasoning depth from $0$ to $6$.
    BERT trained on RP achieves almost perfect accuracy on its test set; however the accuracy drops significantly when it's tested on LP (vice versa).}
    \label{table:RP_LP}
\end{table}

\begin{table}[]
    \setlength{\tabcolsep}{4pt}
    \centering
    {
    \resizebox{\linewidth}{!}{
    {\footnotesize
    \begin{tabular}{ c|c c c c c c c c } 
        \toprule
        Test & $0$ & $1$ & $2$ & $3$ & $4$ & $5$ & $6$  \\ 
        \midrule
         RP\&LP & 99.9 & 99.9 & 99.8 & 99.4 & 98.8 & 98.1 & 95.6 \\  
         
         LP$^*$ &98.1 & 97.2 & \color{red}{92.5} & \color{red}{80.3} & \color{red}{65.8} & \color{red}{55.6} & \color{red}{55.2}\\

        \bottomrule
    \end{tabular}
    }
    }}
    \caption{BERT trained on a mixture over RP and LP fails on LP$^{*}$, a test set that slightly differs from LP.}
    \label{table:RP_LP_2}
\end{table}

\paragraph{Discussion.}
The experiments above reveal a pattern of generalization failure: if we train the model on one data distribution, it fails almost inevitably on a different distribution. In other words, the model seems to be emulating an incorrect ``reasoning function'' specific to its training distribution.

\section{BERT Learns Statistical Features}
\label{sec:story2}
To this point, we have shown that a BERT model achieving high in-distribution accuracy did not learn the correct reasoning function. In this section, we seek to provide an explanation for this peculiar generalization failure.
Our analysis suggests that for the task of logical reasoning, even the simplest statistics of the example can give away significant information about the label, which we denote as \emph{statistical features}. 
Such statistical features are \textbf{inherent} to the task of logical reasoning rather than a problem with specific datasets. 
When BERT is trained on data with statistical features, it tends to make predictions based on such features rather than learning to emulate the correct reasoning function; thus, BERT fails to generalize to the whole problem space. 
However, unlike the shallow shortcuts found in other typical NLP tasks, such statistical features can be countless and extremely complicated, and thus very difficult to be removed from training data. 
\begin{figure}
     \centering
     \begin{subfigure}[b]{\linewidth}
         \centering
         \includegraphics[width=1.0\linewidth]{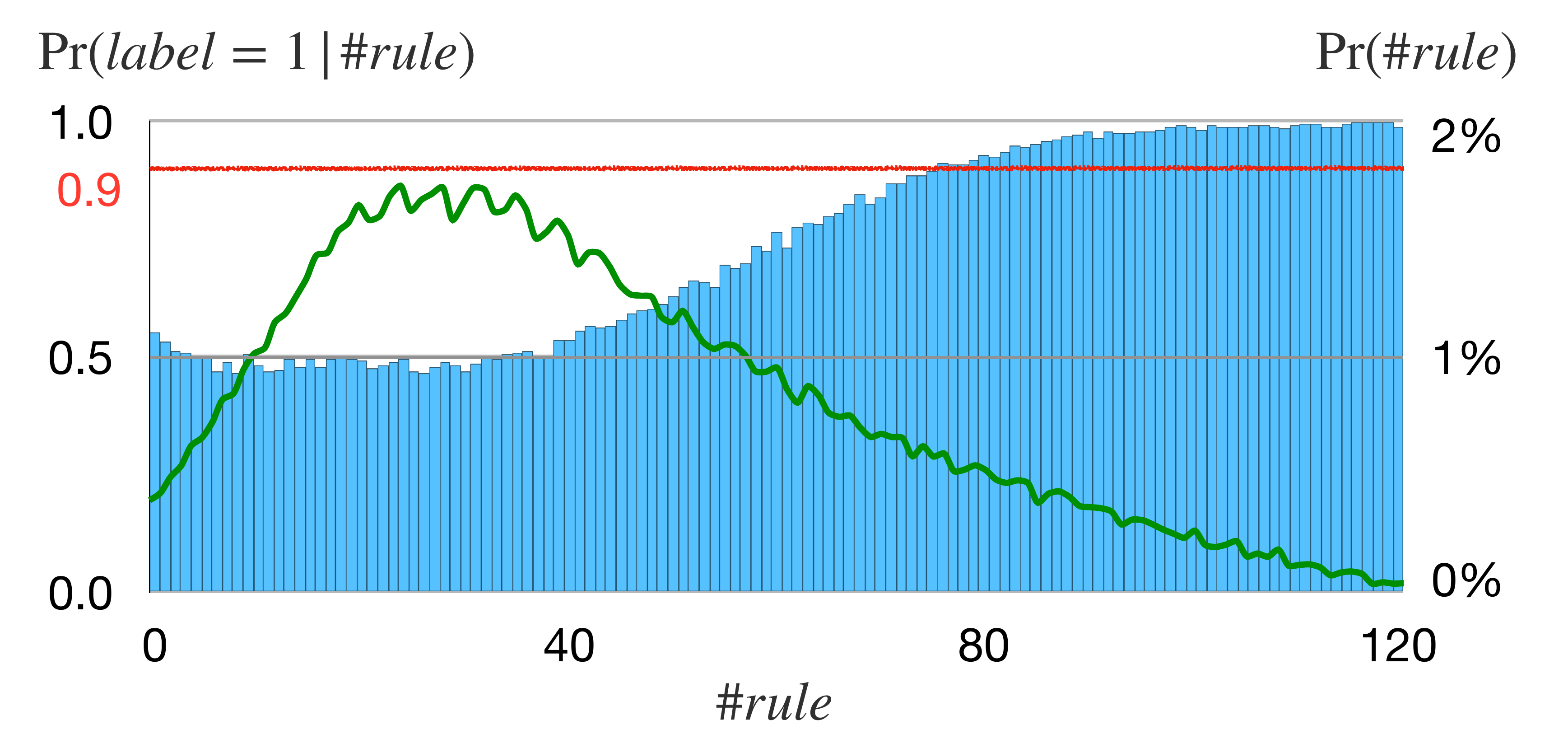}
         \caption{RP: ${\Pr}(\text{label} = 1 \given \text{\#rule}) > 0.5$ for \#rule~$ > 40$.}
         \label{fig:RP_dist}
     \end{subfigure}
     \hfill \\
     \hfill \\
     \begin{subfigure}[b]{\linewidth}
         \centering
         \includegraphics[width=1.0\linewidth]{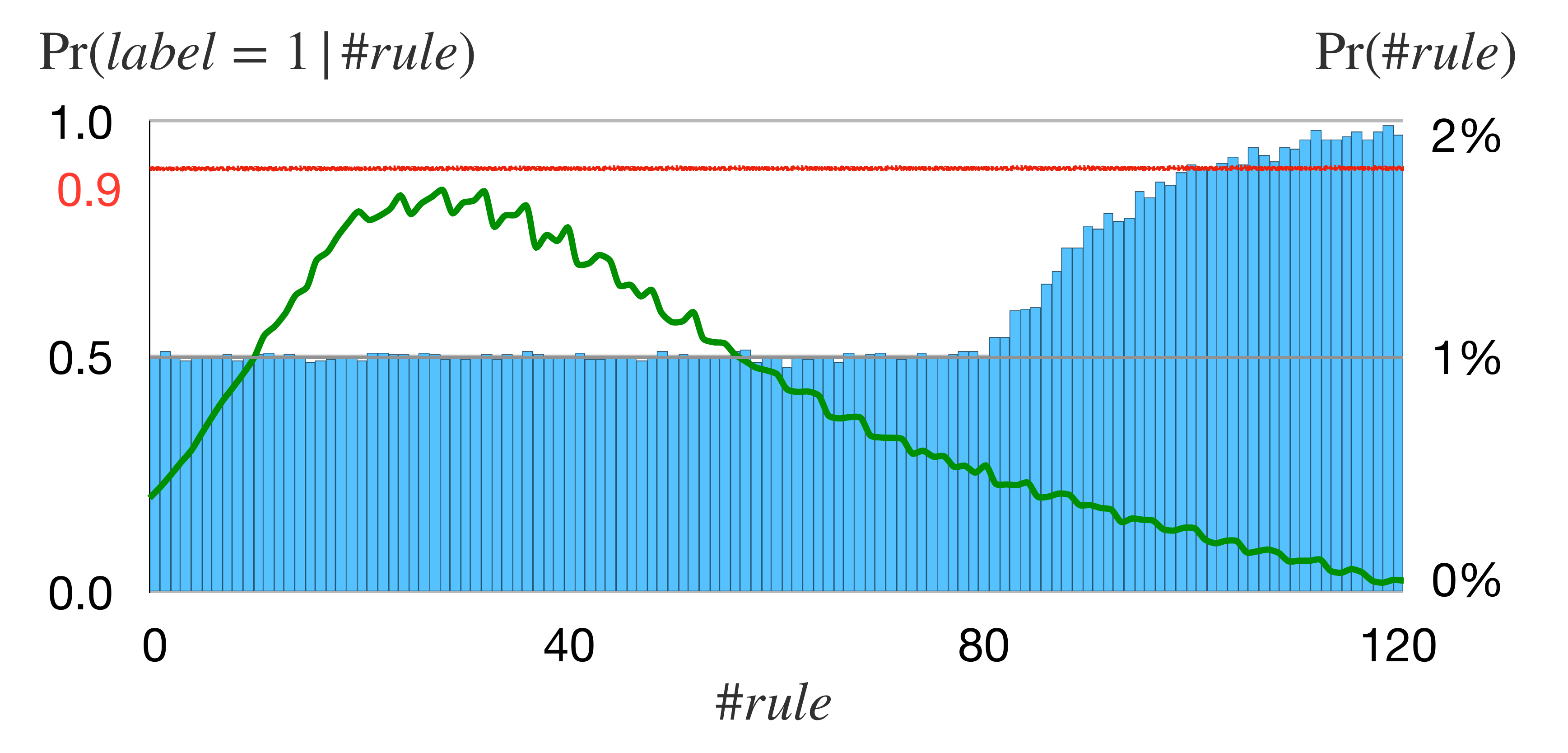}
         \caption{RP\_balance: ${\Pr}(\text{label} = 1 \given \text{\#rule}) \approx 0.5$ for $\text{\#rule} \leq 80$.}
         \label{fig:RP_balance_dist}
     \end{subfigure}
        \caption{$\Pr(\text{label} = 1 \given \text{\#rule})$ (the blue columns) and $\Pr(\text{\#rule})$ (the green curves) for RP and RP\_balance, respectively. After removing \#rule as a statistical feature (RP\_balance), $\Pr(\text{label} = 1 \given \text{\#rule})$ approaches $0.5$ for $\#rule \leq 80$ while $\Pr(\text{\#rule})$ does not change.}
        \label{fig:RP_stats}
\end{figure}

\subsection{Statistical Features Inherently Exists} 
\label{subsec:statistical_feature_exists}
\paragraph{What is a statistical feature?} If a certain statistic of an example has a strong correlation with its label, we call it a \emph{statistical feature}. 

As an illustrating example, we consider the number of rules in a reasoning problem~(\#rule). As shown in Figure~\ref{fig:RP_dist}, the \#rule for reasoning problems in RP exhibit a strong correlation with their labels: when $\text{\#rule} > 40$, the number of positive examples exceeds $50\%$ by large margins; formally, $\Pr_{e \sim \text{RP}}(\text{label(e)} = 1 \given \text{\#rule}(e) = x) > 0.5$ for $x > 40$, which makes it possible for the model to guess the label of an example with relatively high accuracy by only using its \#rule. Hence, we call \#rule a statistical feature for the dataset RP.

\paragraph{Statistical features are inherent to logical reasoning problems.} Continuing with our example, we show that \#rule \emph{inherently} exists as a statistical feature for logical reasoning problems in general; that is, it is not specific to the RP dataset. Consider the following property about logical entailment:
\begin{property}[Monotonicity of entailment] Any additional facts and rules can be freely added to the hypothesis of any proven fact.
\end{property}
It follows that, intuitively, given a fixed set of predicates and facts, any predicate is more likely to be proved when more rules are given, that is, $\Pr_{}(\text{label(e)} = 1 \given \text{\#rule}(e) = x)$ should increase (roughly) monotonically as $x$ increases. Since this intuition assumes nothing about data distributions, it follows that such statistical patterns should naturally exist in any dataset that is not adversarially constructed. In addition to RP, we also verify that both LP and the uniform distribution exhibit similar statistical patterns, which we refer readers to Appendix for further details.

\begin{figure}
     \centering
     \includegraphics[width=1.0\linewidth]{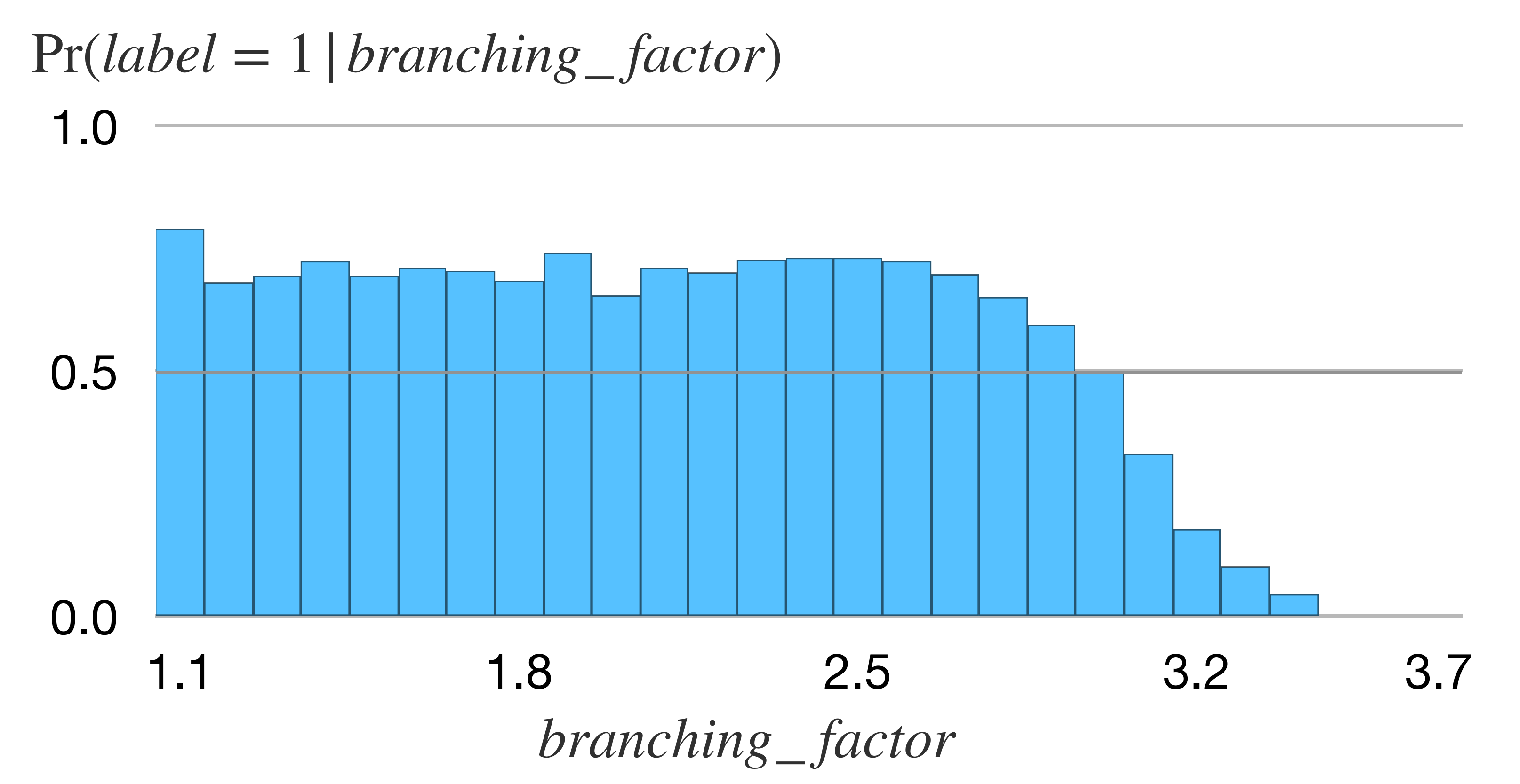}
     \caption{For RP, ${\Pr}(\text{label} = 1 \given \text{branching\_factor})$ decreases as \text{branching\_factor} increases.}
     \label{fig:RP_branching_factor}
\end{figure}

\paragraph{Statistical features are countless.} 
In addition to \#rule, numerous statistical features potentially exist.
For example, as facts can be seen as special form of rules, it follows from previous argument that \#fact is also positively correlated with labels. Statistical features can be more complicated than just \#rule or \#fact. For example, the average number of predicates in rules of a reasoning problem can also leak information about its label. Note that the right-hand side of a rule is only proved if all predicates on its left-hand side are proved. Then, it is immediate that rules of the form $A, B, C \rightarrow D$ are less likely to be ``activated'' than rules of the form $A \rightarrow D$.
Following this intuition, we can define the following statistic: for an example $e$, let 
\begin{align*}
&\text{branching\_factor}(e) \\
&\quad:= \frac{\text{\#fact}(e) + \sum_{\text{rule } \in e} \text{length of rule}}{ \text{\#fact}(e) + \text{\#rule}(e)}.
\end{align*}
In this definition, we are computing the average number of predicates in the rules, where facts are treated as rules with one predicate.\footnote{Branching\_factor: with more predicates on the left-hand side of the rules, the proof tree has more branches.} Our intuition suggests that the larger the branching\_factor, the less likely an example will be positive; we verify that this intuition holds for RP, as shown in Figure~\ref{fig:RP_branching_factor}. Just like \#rule, we observe that branching\_factor is also a statistical feature for LP and the uniform distribution; see details in Appendix.

Now we have shown that, though there are simple statistical features like \#rule, some~(e.g. branching\_factor) can be less intuitive to call to mind; in light of this, it is not hard to imagine that some statistical features can be so complex that they cannot even be manually constructed by humans.
In particular, statistical features can also be \emph{compositional}: one can define a \emph{joint} statistical feature by combining multiple ones (e.g., branching\_factor and \#rule), which further adds to the complexity. Thus, it is infeasible to identify all statistical features.

\begin{table}[t]
    \setlength{\tabcolsep}{4pt}
    \centering
    {\footnotesize
    \resizebox{\linewidth}{!}{
    \begin{tabular}{cc| c c c c c c c } 
        \toprule
        Train & Test & $0$ & $1$ & $2$ & $3$ & $4$ & $5$ & $6$  \\ 
        \midrule
        \multirow{3}{*}{RP\_b} 
        & RP & 99.8 & 99.7 & 99.7 & 99.4 & 98.5 & 98.1 & 97.0 \\
        & RP\_b & 99.4 & 99.6 & 99.2 & 98.7 & 97.8 & 96.1 & 94.4 \\
        & LP & 99.6 & 99.6 & 99.6 & 97.6 & \textcolor{red}{93.1} & \textcolor{red}{81.3} & \textcolor{red}{68.1} \\
        \midrule
        \multirow{3}{*}{RP}  & RP & 99.9 & 99.8 & 99.7 & 99.3 & 98.3 & 97.5 & 95.5  \\
        & RP\_b & 99.0 & 99.3 & 98.5 & 97.5 & 96.7 & \textcolor{red}{93.5} & \textcolor{red}{88.3} \\
        & LP & 99.8 & 99.8 & 99.3 & 96.0 & \textcolor{red}{90.4} & \textcolor{red}{75.0} & \textcolor{red}{57.3} \\
        \bottomrule
    \end{tabular}
    }}
    \caption{The model trained on RP performs worse on RP\_balance (RP\_b). This indicates that the model is using the statistical feature \#rule to make predictions.}
    \label{table:RP_balance}
\end{table}

\subsection{Statistical Features Inhibit Model Generalization}
\label{subsec:statistical_feature_generalization}
Having verified that statistical features inherently exist for logical reasoning problems, in this section we study how they affect the model behavior.
We show that (1) when statistical features are presented in training distributions, BERT tends to utilize them to make predictions; (2) after removing \textbf{one} statistical feature from training data, the model generalizes better.
It follows that statistical features can hinder the model from learning the correct reasoning function, explaining the generalization failure we observed in Section~\ref{sec:story1}.

\paragraph{Example: removing one statistical feature.}
We use \#rule as an example to illustrate how to remove statistical features from a training dataset~$\mcal{D}$; in particular, there are three criteria that we need to satisfy: (1) label is balanced for the feature; (2) the marginal distribution of the feature remains unchanged; (3) the dataset size remains unchanged.

Formally, our first goal is to sample $\mcal{D}^{\prime} \subset \mcal{D}$ such that, for all $x$:
\begin{align*}
\label{eq:bal}
{\Pr}_{e \sim \mcal{D}^{\prime}}(\text{label}(e) = 1 \given \text{\#rule}(e) = x) = 0.5
\end{align*}
Intuitively, this equation says that on $\mcal{D}^{\prime}$, one cannot do better than 50\% by only looking at the \#rule of an example. Specifically, for all possible values of $x$, if $\Pr_{e \sim \mcal{D}}(\text{label}(e) \text{=} 1 \given \text{\#rule}(e) \text{=} x) \text{ > } 0.5$, we drop some positive examples with $\text{\#rule} \text{ = } x$ from $\mcal{D}$; otherwise, we drop some negative examples. 

However, we would not meet the second criterion by naively dropping the minimum number of examples; consider the following statistics for RP:
\vspace{-4pt}
\begin{center}
\resizebox{\linewidth}{!}{
{\footnotesize
\begin{tabular}{c|c|c}
    \toprule
    \multirow{2}{*}{\#rule} &  before drop  &  after drop \\ 
    & \#examples / positive \% & \#examples / positive \% \\
    \midrule
    38 & 6860 / 49.9\% & 6822 / 50.0\% \\ 
    80 & 2322 / 92.7\%  & \textcolor{red}{339} / 50.0\% \\
    \bottomrule
\end{tabular}}}
\end{center}
As shown in the table, if we were to naively drop the minimum number examples from RP such that Equation~1 is satisfied, we will be left with only 339 examples with {\#rule~=~80}, where the number~(6822) of examples with {\#rule~=~38} remains unchanged. This could be a serious issue in terms of dataset \emph{coverage}: examples with some particular \#rule will dominate $\mcal{D}^{\prime}$ and there will not be enough examples for other \#rule. Recall that this is also the reason we choose RP/LP over uniform sampling to generate our datasets~(Sec.~\ref{subsec:RP_LP}).
Hence, we also need to make sure that as we remove statistical features from $\mcal{D}$, their marginal distributions in $\mcal{D}^{\prime}$ stay close to $\mcal{D}$:
\begin{align*}
{\Pr}_{e \sim \mcal{D}^{\prime}}(\text{\#rule}(e)) = {\Pr}_{e \sim \mcal{D}}(\text{\#rule}(e)).
\end{align*}
In this way, $\mcal{D}^{\prime}$'s coverage of examples with different \#rule remains the same as $\mcal{D}$. 

When both criteria (1) and (2) are satisfied, the size of $\mcal{D}^{\prime}$ will be \emph{much smaller} than $\mcal{D}$ and the ratio $k = |\mcal{D}|/|\mcal{D}^{\prime}|$ can be estimated from $\min_{x}\Pr_{e \sim \mcal{D}}(\text{label}(e) \text{=} 1 \given \text{\#rule}(e) \text{=} x)$. Hence, to make sure that criterion (3) is met, that is the size of $\mcal{D}^{\prime}$ is the same as $\mcal{D}$, we need to pre-sample $k \times \mcal{D}$ and obtain $\mcal{D}^{\prime}$ by down-sampling.

Following this approach, by down-sampling from $k \times \text{RP}$, we construct RP\_balance, where \#rule is no longer a statistical feature. A rough estimation shows that if we were to balance $\Pr_{e \sim \text{RP}}(\text{label}(e) \text{ = } 1 \given \text{\#rule}(e) \text{ = } x)$ for $x$ up to $110$, the ratio $k > 100$, that is,
we need to spend over 100x running time (200 hours on a 40-core CPU) to pre-sample roughly 56 million examples; the computational cost would be even more expensive if we want to completely remove \#rule as a statistical feature. Hence, we only balance this conditional probability for $0 \leq x \leq 80$, which takes 10x running time (20 hours on a 40-core CPU) to pre-sample 5.6 million examples. This would not be a major problem as 90\% of the examples in RP have \#rule $\leq 80$.
We train the BERT model on RP\_balance, and the results are reported in Table~\ref{table:RP_balance}.

\paragraph{BERT uses statistical features to make predictions.}
As shown in Table~\ref{table:RP_balance}, BERT trained on RP shows large performance drop when tested on RP\_balance, while BERT trained on RP\_balance shows even better performance on RP than RP-trained BERT. 
Since RP\_balance is down-sampled from RP, the accuracy drop from RP to RP\_balance can only be explained by that BERT trained on RP is using \#rule to make predictions. 

\paragraph{Removing statistical features helps generalization.}
As shown in Table~\ref{table:RP_balance}, compared to RP-trained BERT, BERT trained on RP\_balance achieves higher accuracy when tested on LP; in particular, for examples with reasoning depth $6$, the model trained on RP\_balance attains an accuracy of $68.1\%$, approximately $10\%$ higher than the model trained on RP. This is a clear signal that when \#rule is removed as a statistical feature, the model generalizes better, suggesting that statistical features can hinder model generalization.

\paragraph{Statistical features explain the paradox.}
Now we have a good explanation for the paradox: on the first hand, as we have discussed in Section~\ref{subsec:statistical_feature_exists}, statistical features can be arbitrarily complex and powerful neural models can identify and use them to achieve high in-distribution accuracy; on the other hand, since the correlations between statistical features and the labels can change as the data distribution changes~(see Appendix for details), the model that uses them to make predictions does not generalize to out-of-distribution examples. Besides, we notice that though the BERT model seem to be generalizing well for reasoning examples of depth~$ < 3$, it never achieve 100\% accuracy even when tested in-distribution: no matter how strong the statistical features are, they almost never determine the label with 100\% accuracy.

\subsection{On the Dilemma of Removing Statistical Features}
\label{subsec:statistical_feature_remove}
We show that though removing one statistical feature (e.g., \#rule) from training data can benefit model generalization, it is computationally infeasible to jointly remove multiple statistical features. 

Recall that, in the previous section, when we were trying to remove the statistical feature \#rule from RP, we could only afford to remove it for 90\% of the examples. The general idea is that if a statistical feature $X$ has a very strong correlation with the label on some dataset $\mcal{D}$, i.e. $\Pr_{e \sim \mcal{D}}(\text{label}(e) \text{ = } 1 \given X(e) \text{ = } x)$ is very close to $1$ or $0$, then we would need to sample a lot of examples to have a balanced set. 

The combination of multiple statistical features can give much stronger signal about the label than the individual ones; thus it is much harder to jointly remove them. As an example, we consider removing three statistical features from RP: \#fact~(f), branching\_factor~(b) and \#rule~(r).
\begin{table}
\begin{center}
{\footnotesize
\begin{tabular}{l|c|c}
    \toprule
    $X$ & Pr(label = 1 | $X$) & k$\times$ \\ \hline
    f = 15  & 0.908 &  5.5 \\ 
    f = 15, b $\in$ [2.65,2.75]  & 0.975 &  20.0 \\ 
    f = 15, b $\in$ [2.65,2.75], r = 58 & 0.991 & 55.6 \\
    \bottomrule
\end{tabular}}
\end{center}
\caption{Jointly removing statistical features is difficult; e.g. second row shows: we need to sample \emph{at least} 20 $\times$ RP to balance Pr(label~=~1 | f = 15, b $\in$ [2.65, 2.75]).}
\label{table:jointly_remove}
\end{table}

As shown in Table~\ref{table:jointly_remove}, as we try to jointly remove more statistical features $X$, $\Pr(\text{label} = 1 | X)$ becomes more unbalanced; in particular, as we try to progressively remove \#fact, branching\_factor and \#rule, the minimum times of examples we need to sample grows roughly exponentially: $5.5 \rightarrow 20.0 \rightarrow 55.6$. Besides, the third column in Table~\ref{table:jointly_remove} only shows some lower-bounds for $k$: we are only considering balancing the conditional probability for \emph{one} particular assignment~(\#fact~=~15, braching\_factor~$\in$~[2.65,2.75], \#rule~=~58); for some other assignments, the conditional probability can be more unbalanced, making it even more difficult to jointly remove them.

\section{Related Work}
    A great proportion of NLP tasks require logical reasoning. Prior work contextualizes the problem of logical reasoning by proposing reasoning-dependent datasets and studies solving the tasks with neural models~\citep{johnson2017clevr,sinha2019clutrr,yu2020reclor,liu2020logiQA,tian2021diagnose}. However, most studies focus on solving a single task, and the datasets either are designed for a specific domain \citep{johnson2017clevr,sinha2019clutrr}, or have confounding factors such as language variance \citep{yu2020reclor}; they can not be used to strictly or comprehensively study the logical reasoning abilities of models. 
    
    Another line studies leveraging deep neural models to solve pure logical problems.
    For examples, SAT \citep{selsam2019learning}, maxSAT \citep{wang2019satnet}, temporal logical problems \citep{hahn2021teaching}, DNF counting \citep{crouse2019improving}, logical reasoning by learning the embedding of logical formula \citep{crouse2019improving, abdelaziz2020experimental} and mathematical problems \citep{saxton2019analysing,lample2020deep}. In this work, we focus on deductive reasoning, which is a general and fundamental reasoning problem. \citet{clark2020ruletaker} conducts a similar study to show that models can be trained to reason over language, while we
    observe the difficulty of learning to reason from data. \citet{xu2019can} studies how well neural models can generalize on different types of reasoning problems from a theoretical perspective.

\section{Conclusion}
In this work, we study whether BERT can be trained to conduct logical reasoning in a confined problem space. Our work shows that though BERT can achieve near-perfect performance on a data distribution that covers the whole space, it always fails to generalize to other distributions that are even just slightly different. We demonstrate that the BERT model has not learned to emulate the correct reasoning function: it is in fact learning statistical features, which inherently exist in logical reasoning problems. We further show that it is computationally infeasible to identify and remove all such statistical features from training data, establishing the difficulty of learning to reason.

\section*{Acknowledgements}
This work is partially supported by a DARPA PTG grant, NSF grants \#IIS-1943641, \#IIS-1956441, \#CCF-1837129, Samsung, CISCO, and a Sloan Fellowship. This work is supported in part by Amazon scholarship. 


\bibliography{anthology,custom}
\bibliographystyle{acl_natbib}

\newpage
\appendix
\onecolumn

\section{Statistical Features in Different Data Distributions}
\begin{figure}[H]
    \begin{subfigure}[t]{0.49\linewidth}
        \centering
        \includegraphics[width=0.9\linewidth]{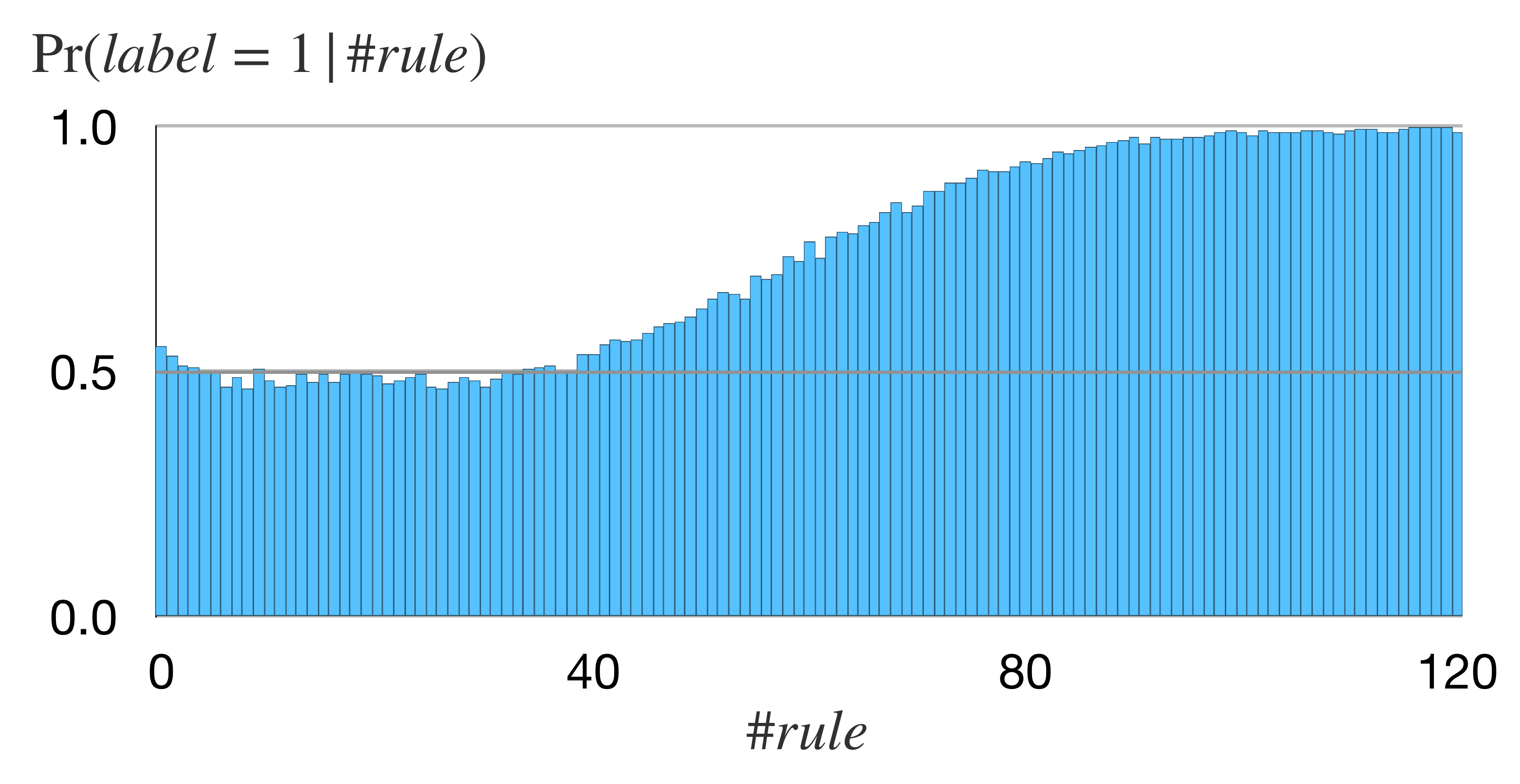}
        \caption{Statistics for examples generated by Rule-Priority (RP).}
    \end{subfigure}
    ~
    \begin{subfigure}[t]{0.49\linewidth}
        \centering
        \includegraphics[width=0.9\linewidth]{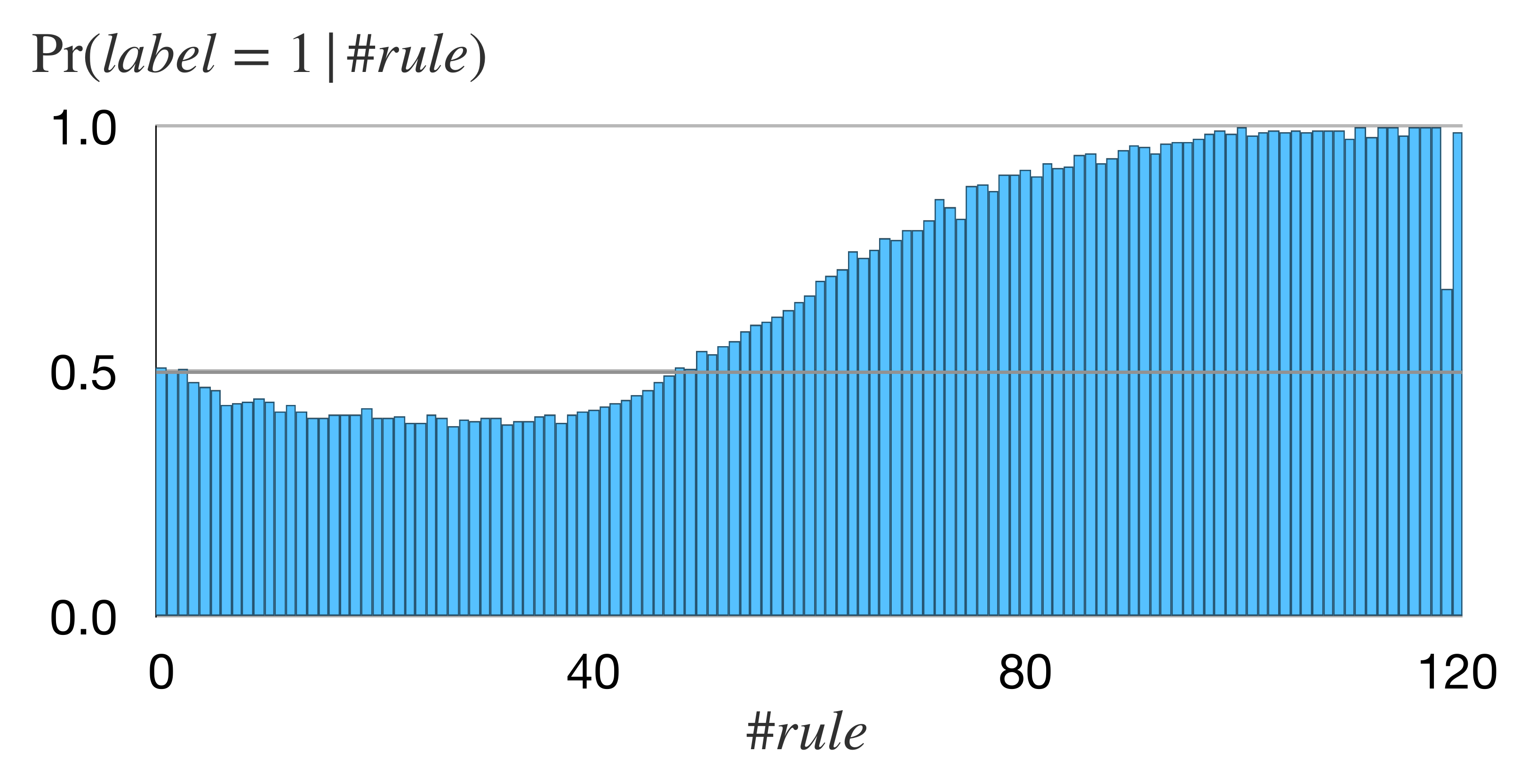}
        \caption{Statistics for examples generated by Label-Priority (LP).}
    \end{subfigure}
    \\
    \hfill \\
    \begin{subfigure}[t]{0.49\linewidth}
        \centering
        \includegraphics[width=0.9\linewidth]{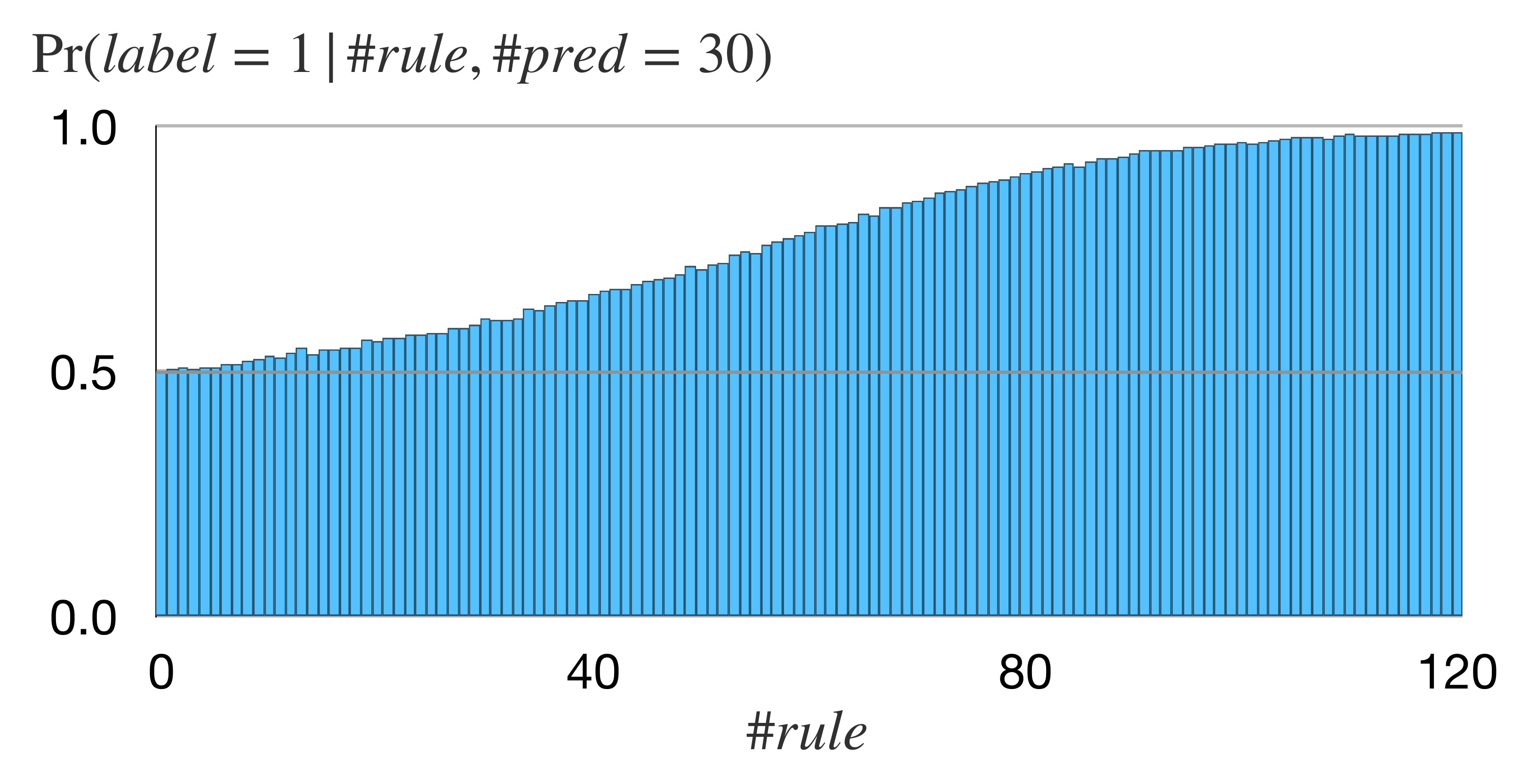}
        \caption{Statistics for examples generated by uniform sampling; we only consider examples with \#pred = 30 as a good-enough approximation: over 99\% of the examples generated by uniform sampling have \#pred = 30.}
    \end{subfigure}
    \caption{\#rule is a statistical feature for RP, LP and the uniform distribution. Even though $\Pr(\text{label} = 1 | \text{\#rule})$ increases as \#rule increases for all three distributions, it follows a slightly different pattern for each distribution; that is to say, the correlation between \#rule and the label changes as the underlying data distribution changes, which explains the generalization failure we observed.}
\end{figure}
\begin{figure}[H]
    \begin{subfigure}[t]{0.49\linewidth}
        \centering
        \includegraphics[width=0.9\linewidth]{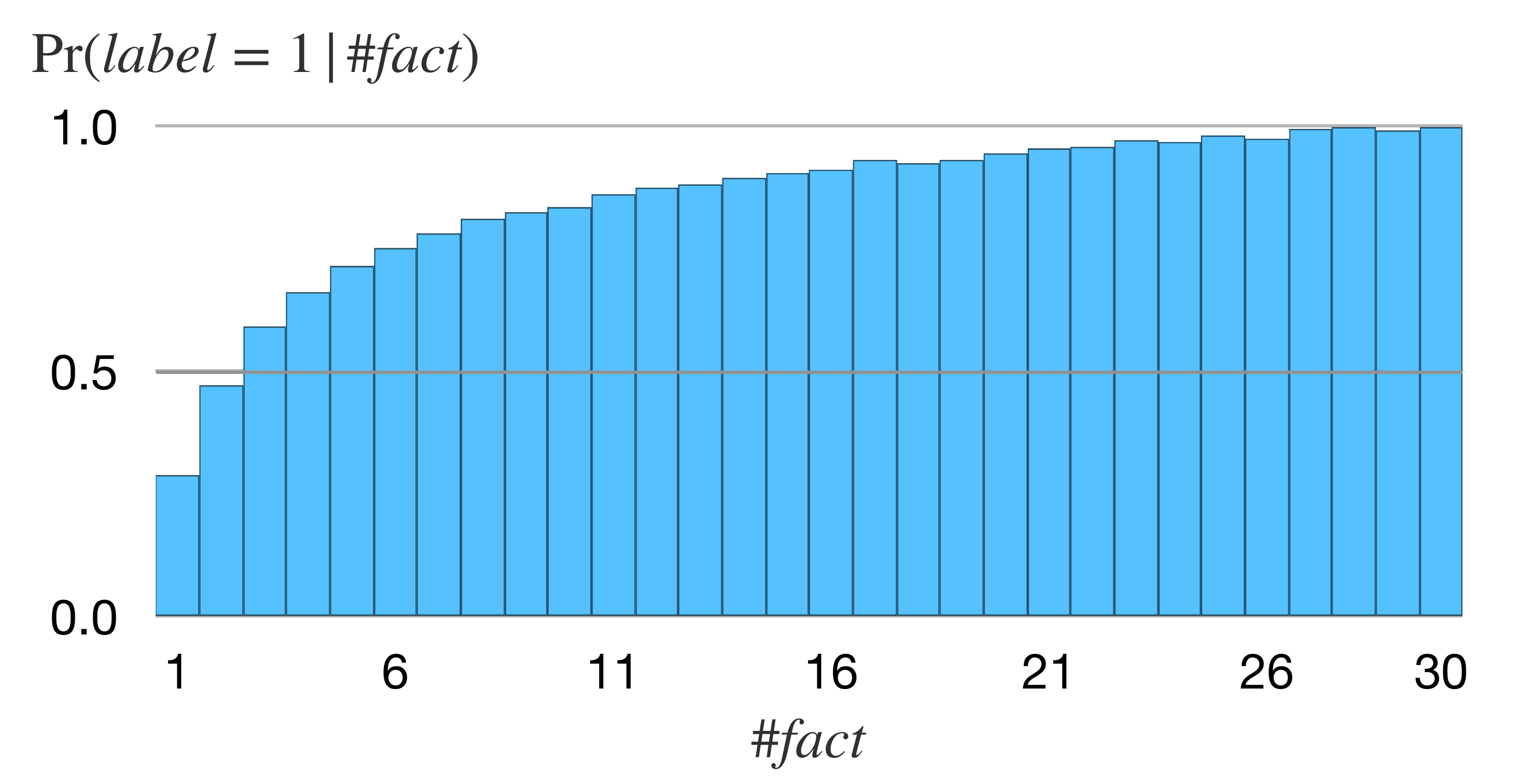}
        \caption{Statistics for examples generated by Rule-Priority (RP).}
    \end{subfigure}
    ~
    \begin{subfigure}[t]{0.49\linewidth}
        \centering
        \includegraphics[width=0.9\linewidth]{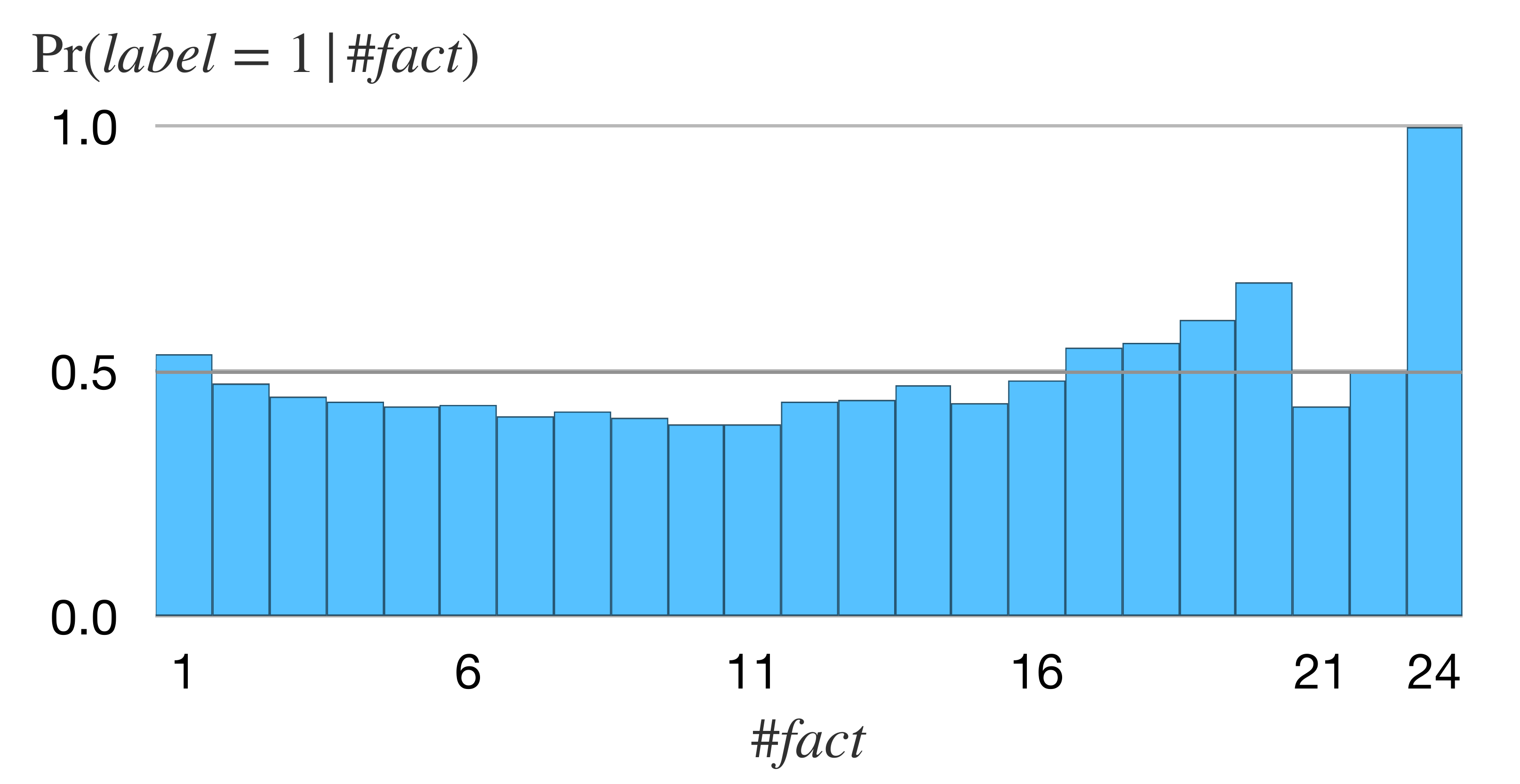}
        \caption{Statistics for examples generated by Label-Priority (LP).}
    \end{subfigure}
    \\
    \hfill \\
    \begin{subfigure}[t]{0.49\linewidth}
        \centering
        \includegraphics[width=0.9\linewidth]{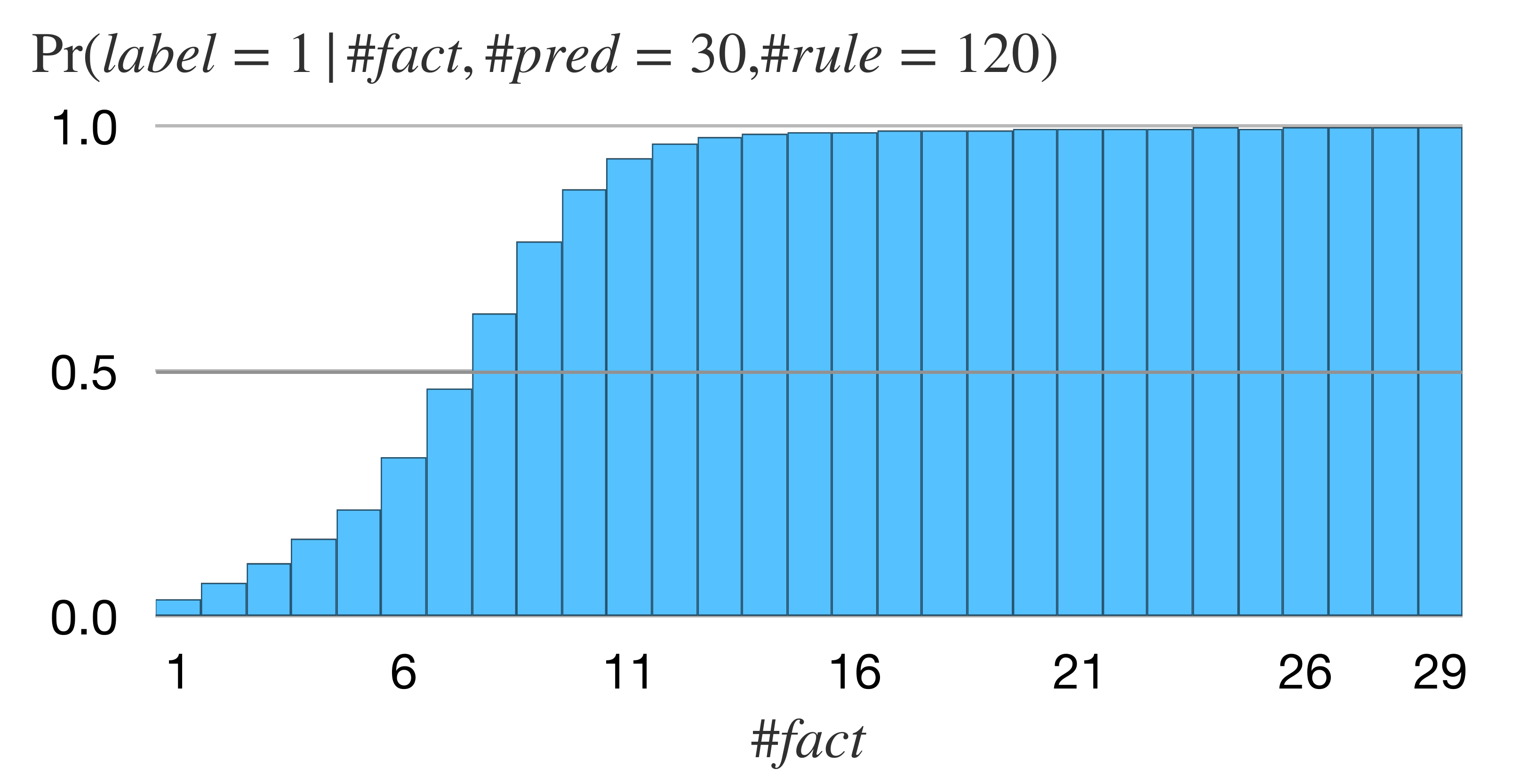}
        \caption{Statistics for examples generated by uniform sampling; we only consider examples with \#pred = 30 and \#rule = 120 as a good-enough approximation: over 99\% of the examples generated by uniform sampling have \#pred~=~30 and \#rule~=~120.}
    \end{subfigure}
    \caption{\#fact is a statistical feature for RP, LP and the uniform distribution.}
    
\end{figure}

\begin{figure}[H]
    \begin{subfigure}[b]{0.49\linewidth}
        \centering
        \includegraphics[width=1.0\linewidth]{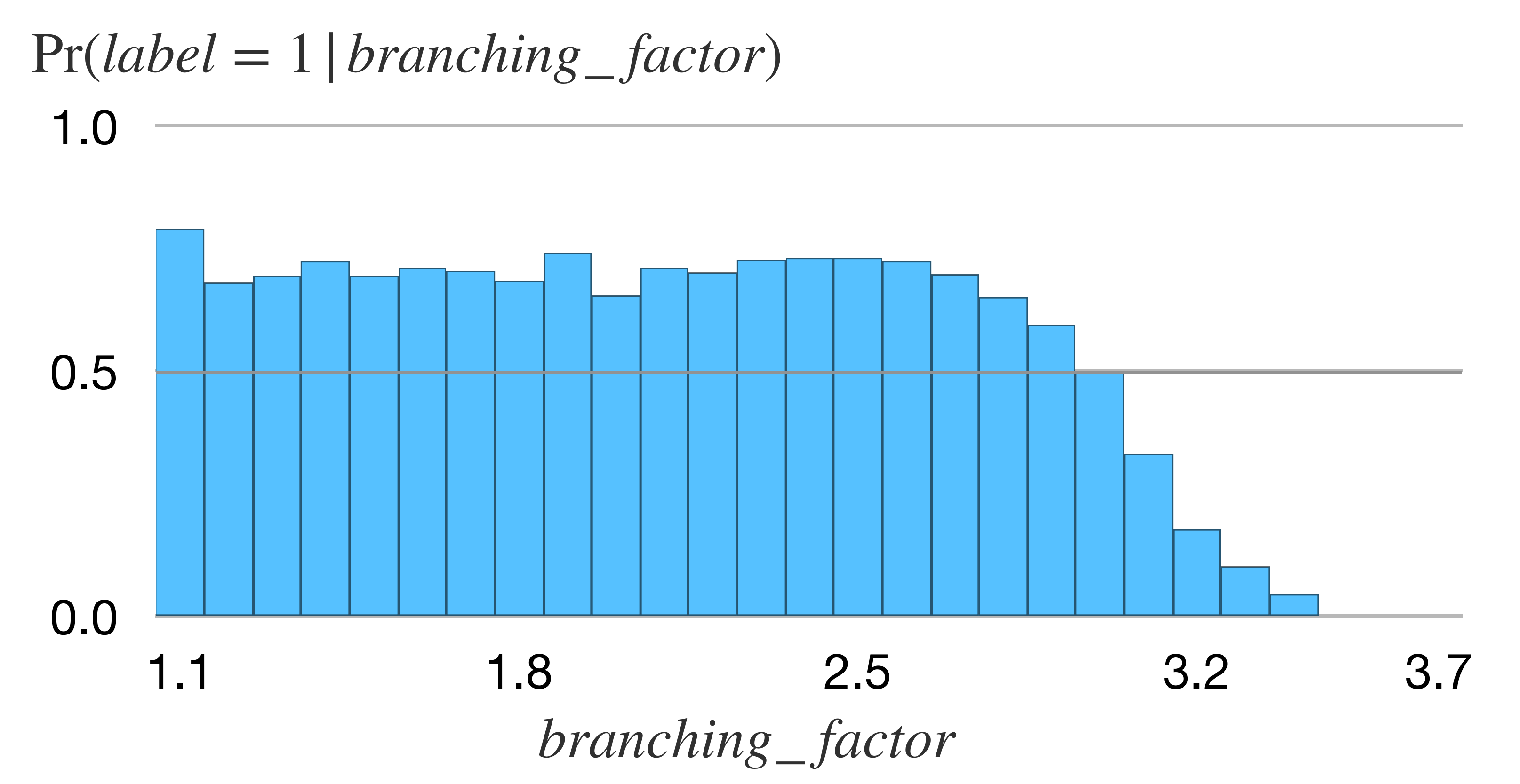}
        \caption{Statistics for examples generated by Rule-Priority (RP).}
    \end{subfigure}
    ~
    \begin{subfigure}[b]{0.49\linewidth}
        \centering
        \includegraphics[width=1.0\linewidth]{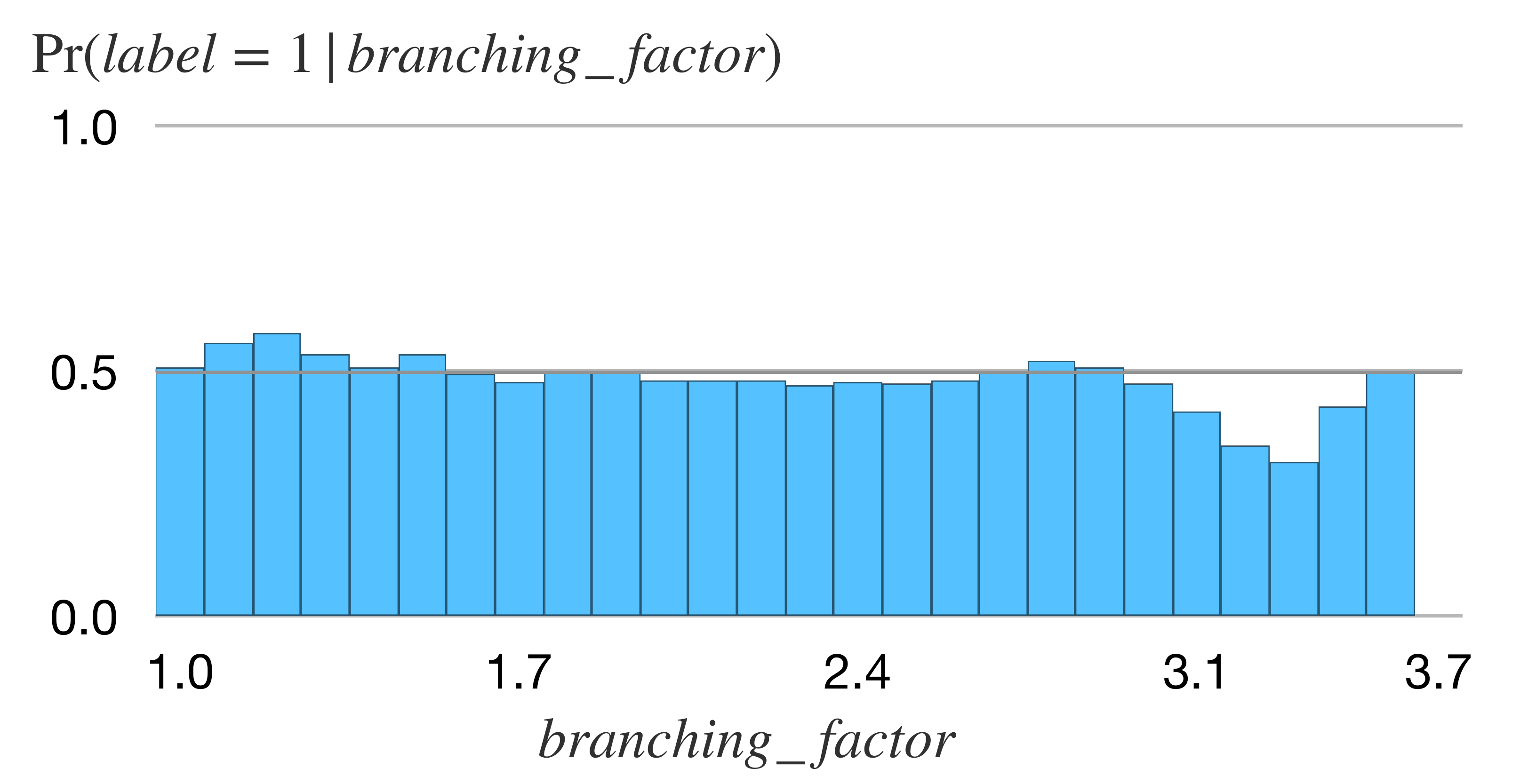}
        \caption{Statistics for examples generated by Label-Priority (LP).}
    \end{subfigure}
    \\
    \hfill \\
    \begin{subfigure}[b]{0.49\linewidth}
        \centering
        \includegraphics[width=1.0\linewidth]{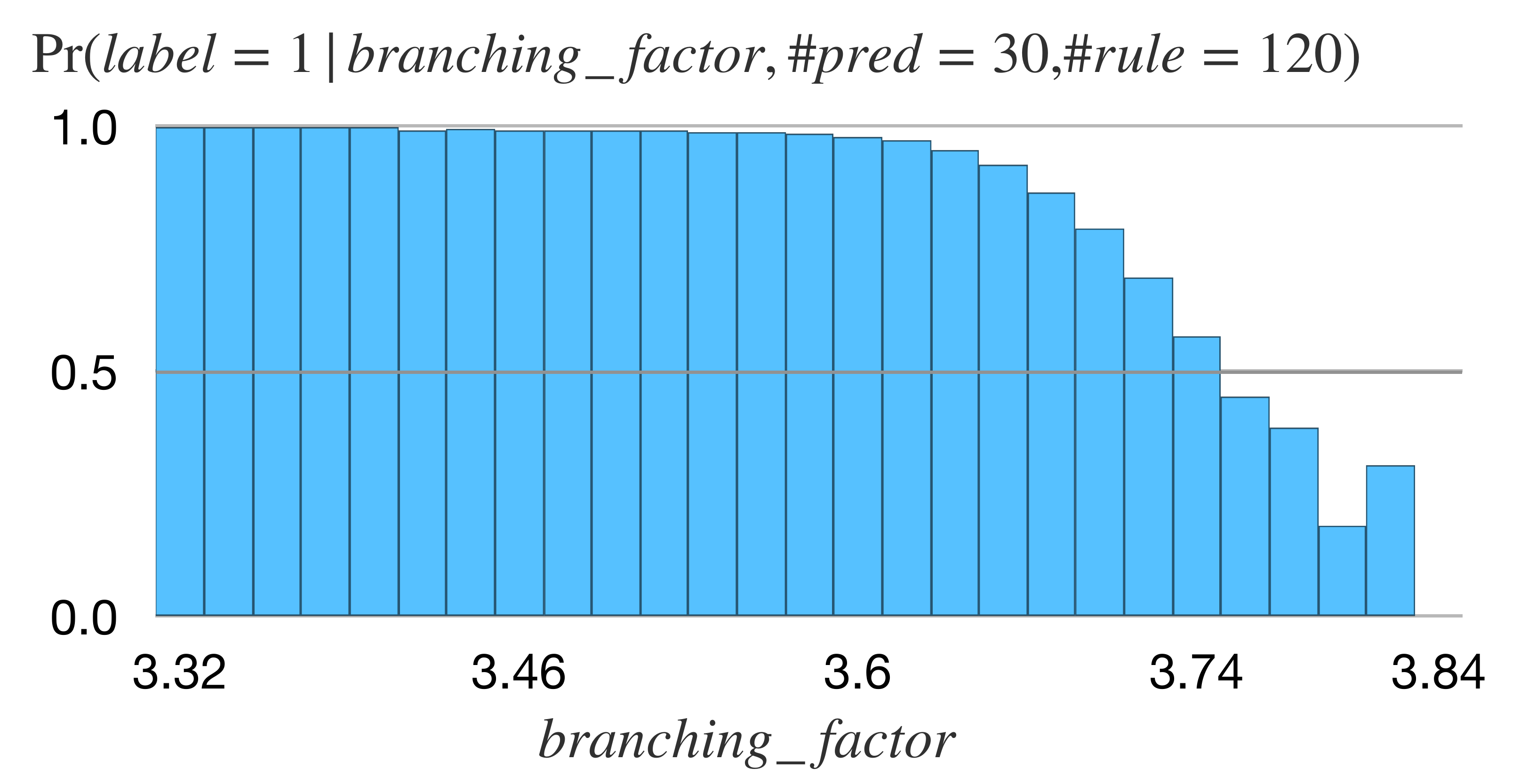}
        \caption{Statistics for examples generated by uniform sampling; we only consider examples with \#pred = 30 and \#rule = 120 as a good-enough approximation: over 99\% of the examples generated by uniform sampling have \#pred~=~30 and \#rule~=~120.}
    \end{subfigure}
    \caption{branching\_factor is a statistical feature for RP, LP and the uniform distribution.}
\end{figure}

\section{Sampling Examples from SimpleLogic}
\label{appendix:LP_RP}
\subsection{Algorithms: Rule-Priority \& Label-Priority}

\begin{figure}[H]
\begin{subfigure}[H]{0.45\linewidth}
\centering
\begin{algorithm}[H]
{\footnotesize
\begin{algorithmic}[1]
\State $pred\_num \sim U[5, 30]$
\State $preds \gets Sample(vocab, pred\_num)$
\State $fact\_num \sim U[1, pred\_num]$
\State $rule\_num \sim U[0, 4 * pred\_num]$
\State $rules$ $\gets$ empty array
\While{size of $rules$ $<$ $rule\_num$}
    \State $body\_num \sim U[1, 3]$
    \State $body \gets Sample(preds, body\_num)$
    \State $head \gets Sample(preds, 1)$
    \If{$tail \not\in body$}
    \State add $body \rightarrow head$ to $rules$
    \EndIf
\EndWhile
\State $fact\_num \sim U[0, pred\_num]$
\State $facts \gets Sample(preds, fact\_num)$
\State $query \gets Sample(preds, 1)$
\State Compute $label$ via forward-chaining.
\State \Return $(facts, rules, query, label)$
\end{algorithmic}
\caption{\emph{Rule-Priority} (RP)}
}
\end{algorithm}
\end{subfigure}
~
\begin{subfigure}[H]{0.45\linewidth}
\centering
\begin{algorithm}[H]
{\footnotesize
\begin{algorithmic}[1]
\State $pred\_num \sim U[5, 30]$
\State $preds \gets Sample(vocab, pred\_num)$
\State $rule\_num \sim U[0, 4 * pred\_num]$
\State set $l \sim U[1, pred\_num / 2]$ and group $preds$ 
\State into $l$ layers
    \For{predicate $p$ in layer $1 \leq i \leq l$}
        \State $q \sim U[0, 1]$
        \State assign label $q$ to predicate $p$
        \If {$i > 1$}
            \State $k \sim U[1, 3]$
            \State $cand \gets$ nodes in layer $i-1$ 
            \State \quad\quad\quad\quad with label $= q$
            \State $body \gets Sample(cand, k)$
            \State add $body \rightarrow p$ to $rules$
        \EndIf
    \EndFor
\While{size of $rules$ $<$ $rule\_num$}
    \State $body\_num \sim U[1, 3]$
    \State $body \gets Sample(preds, body\_num)$
    \State $head \gets Sample(preds, 1)$
    \State add $body \rightarrow tail$ to $rules$ unless $tail$ has label $0$ and 
    \State all predicates in $body$ has label $1$.
\EndWhile
\State $facts \gets$ predicates in layer $1$ with label $=1$
\State $query \gets Sample(preds, 1)$
\State $label \gets$ pre-assigned label for $query$
\State \Return $(facts, rules, query, label)$
\end{algorithmic}
}
\caption{\emph{Label-Priority} (LP)}
\end{algorithm}
\end{subfigure}
\caption{Two sampling algorithms Rule-Priority and Label-Priority. $Sample(X,k)$ returns a random subset from $X$ of size $k$. $U[X,Y]$ denotes the uniform distribution over the integers between $X$ and $Y$.}
\end{figure}

\section{Construction Proof of Theorem 1}
\label{appendix:construction}
We prove theorem \ref{thm:construction} by construction: in N-layer BERT model, we take the first layer as parsing layer, the last layer as output layer and the rest layers as forward chaining reasoning layer. Basically, in the parsing layer we preprocess the natural language input. In forward chaining reasoning layers, the model iteratively broadcast the RHSs to all LHSs, and check the left hand side (LHS) of each rule and update the status of the right hand side (RHS). Here we introduce the general idea of the construction, and we will release the source code for the detailed parameters assignments.

\subsection{Pre-processing Parameters Construction}
\paragraph{Predicate Signature}
For each predicate $P$, we generate its signature $Sign_P$, which is a 60-dimensional unit vector, satisfying that for two different predicates $P_1,P_2,\ Sign_{P_1}\cdot Sign_{P_2}<0.5.$ We can randomly generate those vectors and check until the constraints are satisfied. Empirically it takes no more than 200 trials.

\paragraph{Meaningful Vector}
In parsing layer, we process the natural language inputs as multiple ``meaningful vectors". The meaningful vectors are stored in form of $L_A||L_B||L_C||R||0^{512},$ representing a rule $L_A\wedge L_B\wedge L_C\rightarrow R.$ Each segment $L_A,L_B,L_C,R$ has 64 dimensions, representing a predicate or a always True/False dummy predicate. For each predicate $P$, the first 63 dimensions, denoted as $P^{sign}$, form the signature of the predicate, and the last dimension is a Boolean variable, denoted as $P^v$. The following information is converted into meaningful vectors:
\begin{enumerate}
    \item Rule $LHS\rightarrow RHS:$ if the LHS has less than 3 predicates, we make it up by adding always True dummy predicate(s), and then encode it into meaningful vector, stored in the separating token follows the rule. In addition, for each predicate $P$ in LHS, we encode a dummy meaningful vector as $False\rightarrow P$ and store it in the encoding of $P$. This operation makes sure that every predicate in the input sentence occurs at least once in RHS among all meaningful vectors. We will see the purpose later.
    \item Fact $P$: we represent it by a rule $True \rightarrow P$, and then encode it into meaningful vector and store it in the embedding of the separating token follows the fact.
    \item Query $Q$: we represent it by a rule $Q\rightarrow Q$, encode and store it in the [CLS] token at beginning.
\end{enumerate}

Hence, in the embedding, some positions are encoded by meaningful vectors. For the rest positions, we use zero vectors as their embeddings.
    
\subsection{Forward Chaining Parameters Construction}
Generally, to simulate the forward chaining algorithm, we use the attention process to globally broadcast the true value in RHSs to LHSs, and use the MLP layers to do local inference for each rule from the LHS to the RHS. 

In attention process, for each meaningful vector, the predicates in LHS look to the RHS of others (including itself). If a RHS has the same signature as the current predicate, the boolean value of the RHS is added to the boolean value of the current predicate. Specifically, we construct three heads. We denote $Q_i^{(k)}$ to stand for the query vector of the i-th token of the k-th attention head. For a meaningful vector written as $L_A||L_B||L_C||R||0^{512},$ 
\begin{equation*}
\small
    \begin{aligned}
        Q_i^{(1)}&=L_A^{sign}||\frac 14,Q_i^{(2)}=L_B^{sign}||\frac 14,Q_i^{(1)}=L_C^{sign}||\frac 14\\
        K_i^{(1)}&=\beta R,K_i^{(2)}=\beta R,K_i^{(3)}=\beta R\\   
        V_i^{(1)}&=0^{63}||R^v,V_i^{(2)}=0^{63}||R^v,V_i^{(3)}=0^{63}||R^v.\\  
    \end{aligned}
\end{equation*}
Here $\beta$ is a pre-defined constant. The attention weight to a different predicate is at most $\frac{3\beta}{4}$, while the attention weight to the same predicate is at least $\beta$, and the predicate with positive boolean value has even larger ($\frac{5\beta}{4}$) attention weight. Thus, with a large enough constant $\beta$, we are able to make the attention distribution peaky. Theoretically, when $\beta>300\ln 10,$ we can guarantee that the attention result
$$Attention(Q,K,V)=softmax\left(\frac{QK^T}{\sqrt{d_k}}\right)V$$
satisfies that the value is in the range of $[0.8, 1.0]$ if the predicate on LHS is boardcasted by some RHS with true value, otherwise it is in the range of $[0, 0.2].$

This attention results are added to the original vectors by the skipped connection. After that, we use the two-layer MLP to do the local inference in each meaningful vector. Specifically, we set
    \begin{align*}
        10[&ReLU(L_A^v+L_B^v+L_C^v-2.3)\\
        -&ReLU(L_A^v+L_B^v+L_C^v-2.4)]
    \end{align*}
as the updated $R^v$. Thus, $R^v=1$ if and only if all the boolean values in LHS are true, otherwise $R^v=0$. We also set $L_A^v,L_B^v,L_C^v$ as $0$ for the next round of inference. 

\subsection{Output Layer Parameters Construction}
In output layer, we take out the Boolean value of the RHS of the meaningful vector in [CLS] token.

\section{Examples from SimpleLogic}
\noindent\fbox{%
    \parbox{\textwidth}{%
Rules:  If messy and hypocritical and lonely, then shiny. If tame, then friendly. If plain and shiny and homely, then nervous. If tender, then hypocritical. If dull and impatient and plain, then tame. If spotless, then perfect. If elegant and tender, then homely. If lonely and inquisitive and plain, then homely. If proud, then quaint. If outrageous and homely and impatient, then messy. If quaint, then outrageous. If elegant and glamorous and ugly, then homely. If perfect and sincere and mean, then ambitious. If spotless and quaint and tame, then messy. If tame and sincere and homely, then elegant. If ambitious, then elegant. If shiny and proud, then combative. If quaint and elegant and nervous, then impatient. If glamorous, then outrageous. If proud, then friendly. If combative and nervous, then outrageous. If outrageous and quaint, then careless. If lonely and plain, then inquisitive. If lonely and ugly and combative, then tame. If friendly, then dull. If lonely, then tame. If tender and plain and lonely, then elegant. If glamorous, then hypocritical. If tame and helpless and impatient, then friendly. If careless and messy, then nervous. If combative and shiny, then inquisitive. If plain and outrageous and ugly, then glamorous. If careless and quaint and spotless, then combative. If homely, then helpless. If ambitious, then proud. If messy and ugly, then inquisitive. If perfect, then proud. If helpless and perfect, then elegant. If perfect, then lonely. If lonely and hypocritical, then perfect. If perfect, then friendly. If tender and messy and ambitious, then quaint. If proud, then mean. If outrageous, then perfect. If nervous, then inquisitive. If hypocritical and homely and nervous, then tender. If friendly and dull and outrageous, then ambitious. If glamorous, then proud. If impatient and nervous, then spotless. If mean and quaint and lonely, then spotless. If glamorous, then careless. If dull and mean, then elegant. If homely, then proud. If inquisitive and plain, then ugly. If tender, then homely. If proud and quaint and lonely, then outrageous. If glamorous and perfect and dull, then messy. If helpless and tame and tender, then proud. If friendly and mean, then helpless. If inquisitive, then spotless. If shiny, then tame. If perfect and quaint, then careless. If careless and nervous and combative, then homely. If outrageous and inquisitive and elegant, then hypocritical. If tender and quaint and perfect, then careless. If mean and friendly and ambitious, then combative.

Facts:  Alice shiny. Alice tender. Alice lonely.

Query: Alice is dull ?

Label: True

Proof Depth:  3

From: RP
    }%
}

\noindent\fbox{%
    \parbox{\textwidth}{%
Rules:  If comfortable, then tense. If nervous, then blushing. If nervous and difficult, then beautiful. If disgusted, then clean. If talkative and aggressive, then light. If versatile and supportive, then beautiful. If aggressive, then different. If glamorous and supportive and pleasant, then inexpensive. If light and outrageous and modern, then pleasant. If blushing, then tense. If beautiful, then clean. If perfect and inexpensive, then comfortable. If modern and different, then supportive. If tense, then glamorous. If talkative and aggressive and perfect, then blushing. If versatile, then outrageous. If tense, then perfect. If modern and perfect and inexpensive, then difficult. If versatile and aggressive, then reserved. If comfortable and versatile, then modern. If pleasant and versatile, then reserved. If clean and tense and difficult, then outrageous. If glamorous and modern, then courageous. If elegant and clean, then perfect. If pleasant, then tense. If versatile and blushing and elegant, then light. If reserved, then clean. If clean and talkative and difficult, then reserved. If light, then courageous. If blushing, then light. If different and beautiful, then modern. If disgusted and talkative, then perfect. If elegant and reserved and talkative, then aggressive. If elegant and courageous, then outrageous. If modern and difficult, then disgusted. If supportive and beautiful, then light. If blushing, then glamorous. If comfortable and modern and glamorous, then blushing. If disgusted and inexpensive and talkative, then difficult. If different and clean and disgusted, then modern. If clean and talkative and light, then supportive. If modern and nervous, then difficult. If talkative and aggressive, then modern. If tense and beautiful, then supportive. If modern and inexpensive and glamorous, then comfortable. If difficult and beautiful and modern, then supportive. If nervous and elegant and aggressive, then modern. If tense, then light. If comfortable and inexpensive and disgusted, then tense. If inexpensive and elegant, then nervous. If nervous, then elegant. If glamorous and pleasant, then elegant. If elegant and outrageous, then pleasant. If aggressive and disgusted and comfortable, then light. If talkative and reserved, then clean. If aggressive and modern and inexpensive, then supportive. If reserved and versatile and glamorous, then modern. If comfortable and pleasant and beautiful, then outrageous. If nervous and different and elegant, then modern. If difficult and perfect and outrageous, then tense. If comfortable and blushing and glamorous, then clean. If disgusted, then inexpensive. If inexpensive and tense, then blushing. If elegant, then aggressive. If inexpensive and versatile, then pleasant. If supportive and tense and beautiful, then disgusted. If glamorous and beautiful, then talkative. If tense and reserved, then beautiful. If different, then pleasant. If glamorous and supportive, then clean.

Facts:  Alice versatile. Alice beautiful. Alice light. Alice glamorous. Alice outrageous. Alice difficult.

Query: Alice is comfortable ?

Label: True

Proof Depth:  6

From: RP
    }%
}

\noindent\fbox{%
    \parbox{\textwidth}{
Rules:  If blushing and disgusted, then fancy. If impatient, then long. If frantic, then long. If blushing and frail, then gifted. If frail and long and fancy, then disgusted. If frantic and helpless, then gifted. If broad-minded and frantic, then blushing. If helpless, then broad-minded. If frantic and disgusted and frail, then blushing. If helpless, then impatient. If blushing, then disgusted. If long and gifted and blushing, then frantic. If frantic, then blushing. If fancy, then impatient. If gifted, then fancy. If frail, then helpless. If blushing and frail, then helpless. If blushing, then gifted. If broad-minded and impatient, then long. If broad-minded and disgusted, then fancy. If impatient and disgusted and long, then broad-minded. If broad-minded, then helpless. If disgusted and gifted, then blushing. If gifted and frantic, then fancy. If frail, then broad-minded. If fancy, then broad-minded. If broad-minded, then helpless. If blushing and disgusted, then fancy. If frantic and blushing and gifted, then frail. If frantic, then disgusted. If disgusted, then fancy. If fancy and helpless, then frantic. If frail and disgusted and helpless, then broad-minded. If frantic, then gifted. If long and fancy, then frantic. If blushing, then gifted. If impatient and helpless and gifted, then frantic. If frail and gifted and impatient, then broad-minded. If helpless, then broad-minded.

Facts:  Alice frail.

Query: Alice is disgusted ?

Label: False

Proof Depth:  3

From: LP
}
}

\noindent\fbox{%
    \parbox{\textwidth}{
Rules:  If frantic and helpful, then victorious. If inquisitive and zealous, then bad-tempered. If busy and vivacious, then condemned. If embarrassed, then rude. If thoughtful and rude and helpful, then zealous. If agreeable, then curious. If witty and perfect and thoughtful, then shiny. If impartial and tense, then fine. If frantic and thoughtful and busy, then embarrassed. If agreeable, then pessimistic. If busy and long and embarrassed, then thoughtful. If long and intellectual and fancy, then enchanting. If perfect and victorious and hurt, then zealous. If inquisitive and hurt, then vivacious. If disgusted and tense, then intellectual. If fine, then busy. If fancy and bad-tempered, then fine. If thoughtful, then long. If victorious and condemned, then hurt. If tense, then fine. If frantic, then enchanting. If victorious, then impartial. If agreeable, then enchanting. If hurt and zealous and inquisitive, then fancy. If curious, then frantic. If helpful and zealous, then intellectual. If busy and curious, then agreeable. If curious, then helpful. If curious and victorious, then pessimistic. If witty and shiny and busy, then perfect. If rude and condemned and victorious, then zealous. If witty and embarrassed, then frantic. If perfect and victorious and enchanting, then fancy. If zealous and witty, then rude. If hurt and curious and condemned, then embarrassed. If victorious and busy and disgusted, then intellectual. If fancy and shiny, then enchanting. If hurt and victorious and agreeable, then curious. If thoughtful and helpful, then disgusted. If fancy and intellectual, then shiny. If frantic and impartial, then embarrassed. If impartial, then thoughtful. If pessimistic, then curious. If condemned, then thoughtful. If enchanting, then witty. If zealous and inquisitive and agreeable, then condemned. If fancy and inquisitive, then bad-tempered. If enchanting and fancy and rude, then curious. If vivacious and condemned, then zealous. If perfect, then impartial. If helpful and embarrassed and frantic, then condemned. If helpful, then perfect. If curious, then embarrassed. If condemned, then enchanting. If fine and intellectual, then shiny. If hurt and agreeable, then victorious. If victorious and condemned and rude, then inquisitive. If fancy, then victorious. If impartial and frantic and curious, then hurt. If fancy and long, then vivacious. If hurt and vivacious, then tense. If witty and vivacious and helpful, then embarrassed. If curious, then hurt. If fancy and rude, then zealous. If impartial and shiny and rude, then tense. If pessimistic, then embarrassed. If disgusted and busy and rude, then long. If witty and embarrassed and victorious, then pessimistic. If curious and agreeable, then vivacious. If embarrassed and hurt, then victorious. If intellectual, then witty.

Facts:  Alice tense. Alice disgusted.

Query: Alice is hurt ?

Label: False

Proof Depth:  6

From: LP
    }}

\end{document}